\documentclass[twoside,11pt]{article}
\usepackage{jmlr2e_tepper}

\graphicspath{{data/}}
\usepackage{tikz}
\usetikzlibrary{arrows}

\usepackage[labelformat=simple]{subcaption}
\usepackage{amsmath, amssymb, mathrsfs}
\usepackage{mathtools}

\newtheorem{note}{Note}

\newcommand*{\defeq}{\stackrel{\text{def}}{=}}

\DeclareMathOperator*{\argmin}{\arg\!\min}
\newcommand{\vect}[1]{{\mathbf{#1}}}
\newcommand{\mat}[1]{{\mathbf{#1}}}
\newcommand{\set}[1]{{\mathcal{#1}}}
\newcommand{\norm}[2]{\left\| #1 \right\|_{#2}}
\newcommand{\transpose}[1]{#1^\mathrm{T}}

\newcommand{\Real}{{\mathbb R}}

\newcommand{\bintail}[3]{{\mathcal{B}} \left( #1,#2;#3\right)}
\newcommand{\eps}{\varepsilon}
\newcommand{\meps}{$\eps$\xspace}

\usepackage{stackengine}
\newcommand{\ubar}[1]{\stackunder[1.2pt]{$#1$}{\rule{.8ex}{.05ex}}}

\usepackage[algo2e,ruled,vlined,linesnumbered]{algorithm2e}
\SetKwComment{Comment}{//\ }{}
\makeatletter
\newcommand{\nosemic}{\renewcommand{\@endalgocfline}{\relax}}% Drop semi-colon ;
\newcommand{\dosemic}{\renewcommand{\@endalgocfline}{\algocf@endline}}% Reinstate semi-colon ;
\makeatother

\usepackage{hyperref}
\usepackage[noabbrev,capitalise,compress]{cleveref}
\crefname{section}{Section}{sections}
\crefname{figure}{Figure}{figures}
\crefname{table}{Table}{tables}
\crefname{equation}{Equation}{equations}
\crefrangelabelformat{equation}{(#3#1#4--#5#2#6)}
\crefname{problem}{Problem}{problems}
\creflabelformat{problem}{(#2#1#3)}
\crefname{algorithm}{Algorithm}{algorithms}
\creflabelformat{algorithm}{#2#1#3}
\crefname{assumption}{Assumption}{assumptions}
\creflabelformat{assumption}{(#2#1#3)}
\crefname{step}{Step}{steps}
\creflabelformat{step}{(#2#1#3)}
\crefname{algoline}{Line}{lines}
\crefname{definition}{Definition}{definitions}

\usepackage{booktabs}
\usepackage{multirow}
\usepackage{tabu}
\usepackage{rotating}
\usepackage{siunitx}

\usepackage{paralist}

\begin{document}
    
\title{%
	Fast L1-NMF for Multiple Parametric Model Estimation%
	\thanks{This work was partially supported by Google, NSF, ONR, NGA, and ARO.}%
	}

\author{%
	\name Mariano Tepper \email mariano.tepper@duke.edu \\
    \name Guillermo Sapiro \email guillermo.sapiro@duke.edu \\
    \addr Department of Electrical and Computer Engineering\\
    Duke University
}

\maketitle

\begin{abstract}%
In this work we introduce a comprehensive algorithmic pipeline for multiple parametric model estimation.
The proposed approach analyzes the information produced by a random sampling algorithm (e.g., RANSAC) from a machine learning/optimization perspective, using a \textit{parameterless} biclustering algorithm based on L1 nonnegative matrix factorization (L1-NMF).
The proposed framework exploits consistent patterns that naturally arise during the RANSAC execution, while explicitly avoiding spurious inconsistencies.
Contrarily to the main trends in the literature, the proposed technique does not impose non-intersecting parametric models.
A new accelerated algorithm to compute L1-NMFs allows to handle medium-sized problems faster while also extending the usability of the algorithm to much larger datasets. This accelerated algorithm has applications in any other context where an L1-NMF is needed, beyond the biclustering approach to parameter estimation here addressed.
We accompany the algorithmic presentation with theoretical foundations and numerous and diverse examples.
\end{abstract}

\begin{keywords}
    Multiple parametric model estimation, robust fitting, RANSAC, nonnegative matrix factorization, biclustering, L1 optimization
\end{keywords}

\section{Introduction}

This paper addresses the problem of fitting multiple instances of a given parametric model to data corrupted by noise and outliers. This is a prevalent problem for example in computer vision, found in a wide range of applications such as finding lines/circles/ellipses in images, homography estimation in stereo vision, motion estimation and segmentation, and the geometric analysis of 3D point clouds. Finding a single instance of a parametric model in a dataset  is a robust fitting problem that is hard on its own; the difficulties are exacerbated when multiple instances might be present in the dataset due to the unavoidable emergence of pseudo-outliers (data points that belong to one structure and are usually outliers to any other structure).
Thus, we face the problem of simultaneous robust estimation of model parameters and attribution of data points to the estimated models. These two problems are intrinsically intertwined. Furthermore, the correct number of model instances is not known in advance.

Formally, the data is a set $\set{X} = \{ \vect{x}_i \}_{i=1}^{m}$ of $m$ objects, described by
\begin{equation}
\set{X} = \left( \bigcup_k \set{X}_k \right) \cup \set{O} \quad \text{with} \quad (\forall k)\ \set{X}_k \cap \set{O} = \emptyset.
\end{equation}
The objects in each subset $\set{X}_k$, which might intersect, are (noisy) measurements that can be described with a parametric model $\mu(\theta_k)$, with parameter vector $\theta_k$. In the following, we say that the objects in $\set{X}_k$ are inliers to the model $\mu(\theta_k)$. We also generally refer to a set of objects as inliers, in the sense that there is a statistically meaningful instance of a parametric model that describes them. The objects in $\set{O}$ cannot be described with such a model and we refer to them as outliers. 

Let us define more formally these intuitive concepts.
A model $\mu$ is defined as the zero level set of a smooth parametric function $f_\mu (\vect{x}; \theta)$,
\begin{equation}
	\mu(\theta) = \{ \vect{x} \in \Real^d ,\ f_\mu (\vect{x}; \theta) = 0 \} ,
	\label{eq:model}
\end{equation}
where $\theta$ is a parameter vector.
We define the error associated with the datum $\vect{x} \in \set{X}$ with respect to the model $\mu(\theta)$ as
\begin{equation}
	\operatorname{e}_\mu (\vect{x}, \theta) = \min_{\vect{x}' \in \mu(\theta)} \operatorname{dist} (\vect{x}, \vect{x}') ,
	\label{eq:pointModelDistance}
\end{equation}
where $\operatorname{dist}$ is an appropriate distance function.
Using this error metric, we define the Consensus Set (CS) of a model as
\begin{equation}
	\set{C}_{\mu} (\set{X}, \theta, \delta) = \{ \vect{x} \in \set{X} ,\ \operatorname{e}_\mu (\vect{x}, \theta) \leq \delta \} ,
	\label{eq:consensusSet}
\end{equation}
where $\delta$ is a threshold that accounts for the measurement noise and/or model mismatch.

Multiple Parametric Model Estimation (MPME) seeks to find the set of inliers-model pairs $\left( \set{X}_k, \theta_k \right)$ from the observed $\set{X}$ such that $\set{X}_k = \set{C}_{\mu} \left( \set{X},\theta_k, \delta \right)$. 
This problem is, by itself, ill-posed. It is standard in the literature to implicitly or explicitly impose a penalty on the number of recovered pairs to render it tractable. We also adopt such a design choice.
Notice that once $\set{X}_k$ is found, the corresponding $\theta_k$ can be estimated for example by simple least-squares regression, i.e.,
\begin{equation}
	\hat{\theta}_k = \argmin_{\theta}  \sum_{\vect{x} \in \set{X}_k} \big[ \operatorname{e}_\mu (\vect{x}, \theta) \big]^2 .
	\label{eq:leastSquares}
\end{equation}

MPME is an important but difficult problem, as standard robust estimators, like RANSAC (RANdom SAmple Consensus) \cite{choi2009,ransac}, are designed to extract a single model.  
Let us then begin by formally explaining how does the RANSAC machinery work, to further illustrate the value and perspective of our proposed MPME formulation.

Let us denote by $b$ the minimum number of elements necessary to uniquely characterize a given parametric model, e.g., $b=2$ for lines and $b=3$ for circles. For example, if we want to discover alignments in a 2D point cloud, the elements are 2D points, models $\mu$ are lines, and $b=2$, since a line is defined by two points. A set of $b$ objects is called a minimal sample set (MSS).
Generically, the random-sample-and-model-generation framework can be described by \cref{algo:random_sample}.
We first randomly sample $n$ MSSs, each generating one model hypothesis. The number $n$ is an overestimation of the number of trials needed to obtain a certain number of ``good'' models~\cite{ransac,toldo08,zuliani2005}.
Then, we compute the CS of each model hypothesis using \cref{eq:consensusSet}. Let $\set{U}$ be the set of all these consensus sets. From this point onwards, different algorithms perform different operations on $\set{U}$. Specifically, RANSAC, the algorithm that introduced this framework,
detects a single model by taking the CS from $\set{U}$ with the largest size, and uses it to estimate $\theta$ from $\set{C}$ as in \cref{eq:leastSquares}.

\begin{algorithm2e}[t]
	\SetKwInOut{Input}{input}
	\SetKwInOut{Output}{output}
	
	\Input{set of objects $\set{X}$, parametric function $f_\mu$, inliers threshold $\delta$.}
	\Output{pool $\set{U}$ of consensus sets.}
	
	$b \gets$ minimum number of elements necessary to uniquely characterize model $\mu$, see Equation~(\ref{eq:model})\;
	
	\ForEach{$j \in \{ 1 \dots n \}$}{
		Select at random a minimal sample set (MSS) $\set{X}_{\textsc{mms}(j)}$ of size $b$ from $\set{X}$\; \label[algoline]{algo:ransac_mss}
		\label[algoline]{algo:random_sample_mss}
		Estimate $\theta_j$ from $\set{X}_{\textsc{mms}(j)}$ by solving $(\forall \vect{x} \in \set{X}_{\textsc{mms}(j)})\ f_\mu (\vect{x}; \theta) = 0$\;
		\label[algoline]{algo:random_sample_estimate}
	}

	$\set{U} \gets \left\{ \set{C}_{\mu} \left( \set{X}, \theta_j, \delta \right) \right\}_{j=1}^{n}$, see \cref{eq:consensusSet}\;
	\label[algoline]{algo:random_sample_all_models}
	
	\caption{Random sampling algorithm}
	\label{algo:random_sample}
\end{algorithm2e}

Applying RANSAC sequentially, removing the inliers from the dataset as each model instance is detected, has been proposed as a solution for multi-model estimation, e.g.,~\cite{rabin2010macransac}. However, this approach is known to be suboptimal~\cite{zuliani2005}. The multiRANSAC algorithm~\cite{zuliani2005} provides a more effective alternative, although the number of models must be known a priori, imposing a very limiting constraint in many applications.
An alternative approach consists of finding density modes in a parameter space. The overall idea is that we can map the data into the parameter space through random sampling, and then seek the modes of the distribution by discretizing the distribution, i.e., using the Randomized Hough Transform~\cite{xu1990}, or by using non-parametric density estimation techniques, like the mean-shift clustering algorithm~\cite{subbarao2006}. These, however, are not intrinsically robust techniques, even if they can be robustified with outliers rejection heuristics~\cite{toldo08}. Moreover, the choice of the parametrization and its discretization are critical, among other important shortcomings~\cite{toldo08}.
The computational cost of these techniques can be very high as well.
From a different perspective, J-linkage~\cite{toldo08}, T-linkage~\cite{magri2014}, and RPA~\cite{magri2015} address the problem by clustering the dataset. We will provide more details about these methods in the next section.

\paragraph{Contributions.}
We have previously introduced a novel framework and perspective to reach consensus in grouping problems by re-framing them as biclustering problems~\cite{Tepper2014consensus}.
In particular, the task of finding/fitting multiple parametric models to a dataset was, for the first time, formally posed as a consensus/biclustering problem.
In this work, we build upon this framework, introducing a complete and comprehensive algorithmic pipeline for multiple parametric model estimation.
The proposed approach preserves and considers all the information produced by \cref{algo:random_sample} (i.e., no averaging/pooling is performed). This information is then analyzed from a machine learning/optimization perspective, using a \textit{parameterless} biclustering algorithm based on nonnegative matrix factorization (NMF).

Contrarily to the main trends in the literature, the proposed modeling does not impose non-intersecting subsets $\set{X}_i$.
Secondly, it exploits consistencies that naturally arise during the RANSAC execution, while explicitly avoiding spurious inconsistencies.
This new formulation conceptually changes the way that the data produced by the popular RANSAC, and related model-candidate generation techniques, is analyzed.

With respect to our previous work \cite[Section~5]{Tepper2014consensus}, the present work includes the following key differences and improvements:
\begin{compactitem}
	\item We have streamlined the pre- and post-processing steps of the biclustering algorithm, simplifying algorithmic choices and unifying the pipeline for different types of parametric models;
	\item No user intervention (i.e., no parameter tuning) is required to find the number of parametric models. During the biclustering process, this number is automatically determined using model selection techniques (i.e., minimum description length);	
	\item We accelerate the biclustering process introducing an accelerated algorithm to solve L1-based nonnegative matrix factorization (L1-NMF) problems. This allows to solve  medium-sized problems faster while also extending the usability of the algorithm to much larger datasets. This contribution exceeds the context of this work, as this algorithm has potential applications in any other context where an L1-NMF is needed (e.g., traffic analysis~\cite{Tepper2015}, eldercare~\cite{Tepper2015}, shadow removal for face detection~\cite{Zhao2016}, video surveillance~\cite{Zhao2016}).
\end{compactitem}
We provide the complete source code of the proposed approach for generating all the examples presented in this work.\footnote{\url{https://github.com/marianotepper/arse}}

\paragraph{Organization.}
The remainder of this paper is organized as follows. In \cref{sec:rse} we present the proposed \emph{biclustering} approach for multiple parametric model estimation. In \cref{sec:arse} we describe the fast version of the biclustering algorithm. We present diverse and extensive experimental results in \cref{sec:results} and finally provide some closing remarks in Section~\ref{sec:conclusions}.

\section{Random Sample Ensemble}
\label{sec:rse}

Let $\mat{X}$ be a matrix. In the following, $(\mat{X})_{ij}$, $(\mat{X})_{:j}$, $(\mat{X})_{i:}$ denote the $(i,j)$-th entry of $\mat{X}$, the $j$-th column of $\mat{X}$, and the $i$-th row of $\mat{X}$, respectively.

The input of the algorithm is a pool $\set{U} = \left\{ \set{C}_{\mu} \left( \set{X}, \theta_j, \delta \right) \right\}_{j=1}^{n}$ of consensus sets (the output of \cref{algo:random_sample}).

\begin{definition}[Preference matrix]
	 We define the $m \times n$ preference matrix $\mat{A}$, whose rows and columns represent respectively the $m=|\set{X}|$ data elements $\{ \vect{x}_i \}_{i=1}^m$ and the $n=|\set{U}|$ consensus sets, as
	 \begin{equation}
		 (\mat{A})_{ij} =
		 \begin{cases}
			 1 & \text{if } \vect{x}_i \in \set{C}_{\mu} \left( \set{X}, \theta_j, \delta \right); \\
			 0 & \text{otherwise.} \\
		 \end{cases}
	 \end{equation}
	 \label[definition]{def:preference_matrix}
\end{definition}

\begin{figure}[t]
	\centering
	
	\begin{tabu} to .8\textwidth {X[.7,c,m] X[c,m]}
		\includegraphics[width=\linewidth]{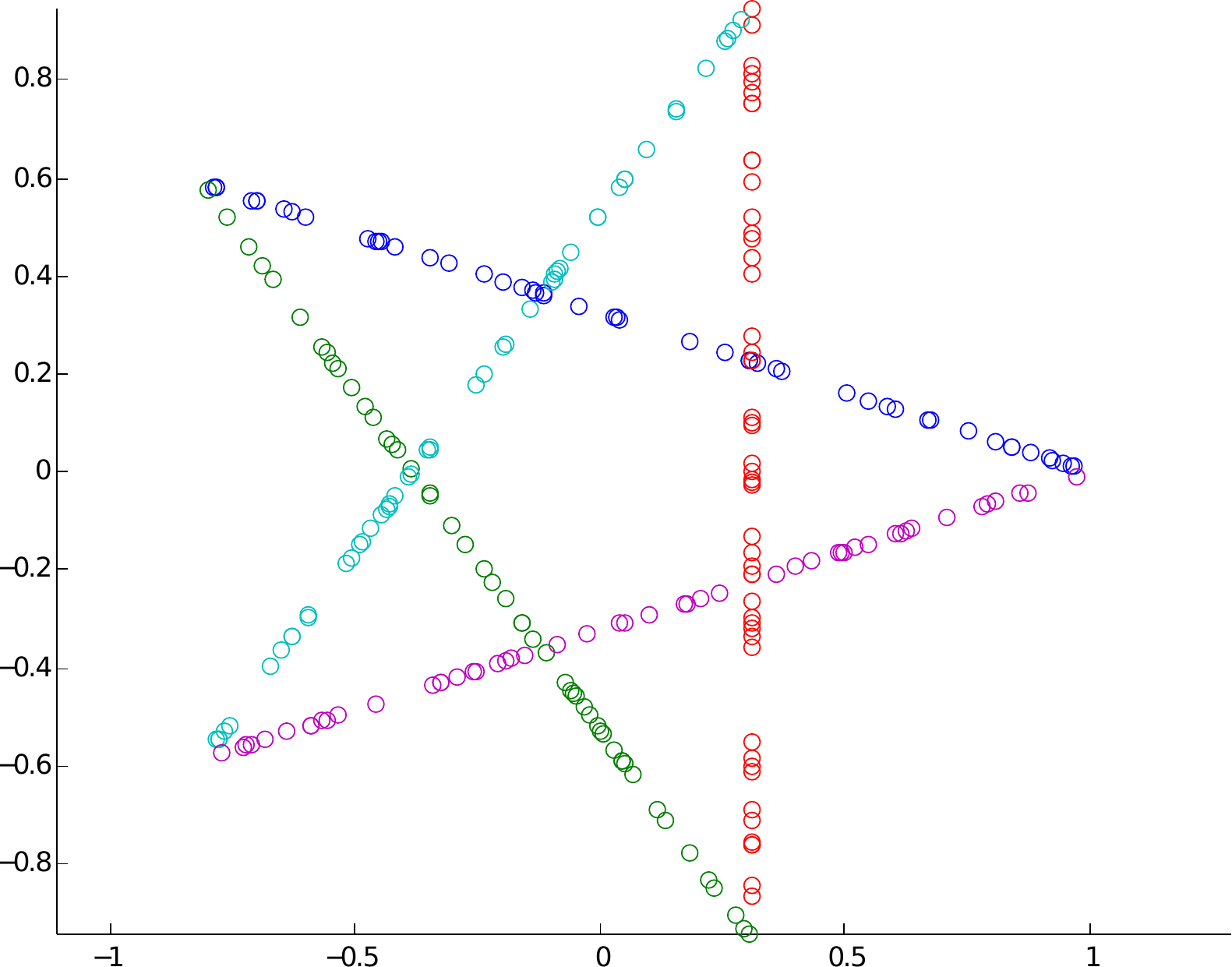} &
		\includegraphics[width=\linewidth]{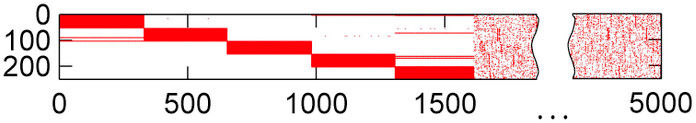}
	\end{tabu}
	
	\caption{Preference matrix example. The data (objects) consist of 250 points on five line segments (models) forming a star. The rows of the preference matrix $\mat{A}$ were reordered (permuted) by group for improved visualization.}
	\label{fig:preferenceMatrix}
\end{figure}

An example of a preference matrix for a synthetic dataset can be seen in \cref{fig:preferenceMatrix}.
Traditionally, in algorithms like RANSAC, the preference matrix is (often implicitly) analyzed column-by-column or row-by-row.
The preference matrix was explicitly introduced in the context of MPME in~\cite{toldo08}.
In the original formulation, the objects in $\set{X}$ were clustered using the rows of $\mat{A}$ as feature vectors, obtaining a powerful state-of-the-art technique called J-linkage. An extension to work with a non-binary (using soft versus hard element-model membership) version of $\mat{A}$ was proposed in~\cite{magri2014}.

An alternative approach involves creating an $m \times m$ co-occurrence matrix, defined as
\begin{equation}
	\mat{B} = \tfrac{1}{n} \sum_{k=1}^{n} \mat{B}_k ,
	\quad \text{where} \quad
	(\mat{B}_k)_{ij} =
	\begin{cases}
	1 & \text{if } x_i, x_j \in \set{C}_{\mu} \left( \set{X}, \theta_k, \delta \right); \\
	0 & \text{otherwise.}
	\end{cases}
	\label{eq:co-occurrence}
\end{equation}
The matrix $\mat{B}$ is widely used in the context of ensemble clustering where it is built from clusters obtained from multiple clustering algorithms, instead of consensus sets. There are, within this research area, many algorithms to analyze $\mat{B}$: from simple techniques such as applying a clustering algorithm to it (e.g., $k$-means, hierarchical or spectral clustering), to more complex matrix factorization techniques~\cite[e.g.,][]{li2007}.
It is important to point out that $\mat{A}$ contains more information than $\mat{B}$. Noticing that
\begin{equation}
\mat{B} = \tfrac{1}{n} \sum_{j=1}^{n} (\mat{A})_{:j} \transpose{(\mat{A})_{:j}} ,
\label{eq:co-occ_preference}
\end{equation}
it becomes clear that the averaging operation implies the loss of critical information.

In the vast majority of matrix factorization/clustering methods, the number of factors/clusters is a critical and hard-to-tune parameter. But this is in fact one of the parameters we are interested in discovering!
In the context of MPME, matrix factorization has been recently applied~\cite{magri2015} to a normalized version of $\mat{B}$ (using soft membership). Provided that the correct number of factors is selected, this method delivers state-of-the-art results.

An additional constraint common to all the aforementioned clustering/factorization approaches is that they provide a segmentation of the dataset elements. This means that these models do not correctly handle intersecting/overlapping models. As we will see next, this limitation is not present in the proposed formulation.

\subsection{Analyzing the preference matrix}
\label{sec:biclustering}

Our approach follows a different conceptual path, simultaneously analyzing the rows and columns of the preference matrix $\mat{A}$. To explain the rationale behind our formulation, let us rewrite \cref{eq:model} for clarity purposes:
\begin{equation*}
\set{X} = \left( \bigcup_{k=1}^{r} \hat{\set{X}}_k \right) \cup \set{O} \quad \text{with} \quad (\forall k)\ \hat{\set{X}}_k \cap \set{O} = \emptyset.
\end{equation*}
Each $\hat{\set{X}}_k$ represents one of the \emph{ground truth} groups that we are seeking to recover.
We assume $(\forall k)\ |\hat{\set{X}}_k| \geq b$, where $b$ is the minimum number of elements necessary to uniquely characterize a given parametric model.

Let us assume that during the execution of \cref{algo:random_sample} we obtained several pure MSSs for some $\hat{\set{X}}_k$, i.e., $\set{J}_k = \{ j \ |\ \set{X}_{\textsc{mms}(j)} \subseteq \hat{\set{X}}_k \}$ (the index $j$ corresponds to the iterations of \cref{algo:random_sample}). Then, for \emph{almost all} $j, j' \in \set{J}_k$ their respective models will be very similar, i.e., $\theta_{j} \approx \theta_{j'}$; these models should also be similar to the model estimated from $\hat{\set{X}}_k$ using least squares. Therefore their consensus sets should be similar, i.e., $\set{C}_{\mu} \left( \set{X}, \theta_j, \delta \right) \approx \set{C}_{\mu} \left( \set{X}, \theta_{j'}, \delta \right) \approx \hat{\set{X}}_k$. From the definition of $\mat{A}$, we can then write that
\begin{equation}
	(\mat{A})_{:\set{J}_k} = \hat{\vect{u}}_{k} \transpose{\vect{1}} + \mat{O}_k,
\end{equation}
where $\hat{\vect{u}}_{k}$ is a binary vector such that $(\forall j \in \set{J}_k)\ (\mat{A})_{:j} \approx \hat{\vect{u}}_{k}$, $\vect{1}$ is the ``all ones'' vector of size $|\set{J}_k|$, and $\mat{O}_k$ accounts for all the errors in this approximation. Finally, as we look to represent all columns of $\mat{A}$, we can rewrite the above equation, adding the corresponding zeros, as $\mat{A} = \hat{\vect{u}}_{k} \transpose{\hat{\vect{v}}_k} + \mat{O}_k$, where $\hat{\vect{v}}_{k}$ is a binary vector of size $n$ such that $(\forall j \in \set{J}_k)\ (\hat{\vect{v}}_{j})_i = 1$ and $\mat{O}_k$ is now of size $m \times n$.

We can then extend this representation to all ground truth groups $\hat{\set{X}}_k$, by writing
\begin{equation}
	\mat{A} = \sum_{k=1}^{r} \hat{\vect{u}}_{k} \transpose{\hat{\vect{v}}_{k}} + \mat{O},
	\label{eq:preference_matrix_approximation}
\end{equation}
where $\hat{\vect{u}}_{k} \in \{0, 1\}^{m}$, $\hat{\vect{v}}_{k} \in \{0, 1\}^{n}$ and again $\mat{O}$ accounts for all errors in the approximation.
Notice that \Cref{eq:preference_matrix_approximation} allows to cast the problem of analyzing the output of \cref{algo:random_sample} in terms of a biclustering problem.

Once the biclusters of $\mat{A}$ have been identified, we can obtain the consensus sets and, from them, estimate the final output models. See \cref{algo:rse} for a full description of this process.
The next question now is: how do we find each bicluster?

\begin{algorithm2e}[t]
	\SetKwInOut{Input}{input}
	\SetKwInOut{Output}{output}
	
	\begin{small}
	\Input{set of objects $\set{X}$, parametric function $f_\mu$, inliers threshold $\delta$.}
	\Output{collection of inliers-model pairs $\{ (\set{C}_t, \theta_t) \}_{t=1}^{T}$.}
	
	Execute \cref{algo:random_sample} to obtain $\set{U}$\;
	Form the preference matrix $\mat{A}$ from $\set{U}$ (\cref{def:preference_matrix})\;
	Obtain $T$ biclusters $\{ (\vect{u}_t, \vect{v}_t) \}_{t=1}^{T}$ from $\mat{A}$ (see \cref{sec:biclustering} for details, note that $T$ is automatically estimated)\;
	\ForEach{$\vect{u}_t$}{
		$\set{C}_t \gets \{ x_i \in \set{X} \ |\ (\vect{u}_t)_i > 0 \}$\;
		Estimate $\theta_t$ from $\set{C}_t$ using least-squares, see \cref{eq:leastSquares}\;
		Re-compute $\set{C}_t$ as $\set{C}_{\mu} \left( \set{X}, \theta_t, \delta \right)$, see \cref{eq:consensusSet}\;
	}
	\end{small}
	
	\caption{Random Sample Ensemble}
	\label{algo:rse}
\end{algorithm2e}

\subsection{L1-NMF}

As usual in the optimization literature, the problem of finding $\hat{\vect{u}}$ and $\hat{\vect{v}}$ in \cref{eq:preference_matrix_approximation} is easier to solve if we soften the binary constraints, imposing nonnegativity instead.
The only missing component to formalize the problem is an appropriate prior for $\mat{O}$. Since the errors are also binary (and thus spurious), a reasonable choice would be to penalize its L1 norm. For convenience, we also drop the binary constraint on $\mat{O}$.
We can thus write the single-bicluster estimation problem as
\begin{equation}
\min_{\vect{u}, \vect{v}} \norm{\mat{A} - \vect{u} \transpose{\vect{v}} - \mat{O}}{F}^2
\quad \text{s.t.} \quad
\vect{u} \geq \vect{0}, \vect{v} \geq \vect{0}, \norm{\mat{O}}{1} \leq \sigma,
\end{equation}
which, for some $\sigma$, is equivalent to our proposed formulation
\begin{equation}
\min_{\vect{u}, \vect{v}} \norm{\mat{A} - \vect{u} \transpose{\vect{v}}}{1}
\quad \text{s.t.} \quad
\vect{u} \geq \vect{0}, \vect{v} \geq \vect{0}.
\tag{L1-NMF}
\label[problem]{eq:l1-nmf}
\end{equation}
We use the L1 norm to cope with the impulsive nature of the errors in $\mat{A}$. This formulation computes a robust median approximation to the preference matrix $\mat{A}$, which carries all the needed information.
Any standard NMF algorithm can be adapted to use the L1 norm and solve~(\ref{eq:l1-nmf}).
The algorithm in \cite[App.~A]{Tepper2014consensus} delivers high-quality solutions at the expense of a higher computational cost.
In this work, for the sake of speed, we use the following procedure:
\begin{compactenum}
	\item Find initializations for $\vect{u}$ and $\vect{v}$ using the iterative re-weighting scheme in \cite{Kong2011}. In our experiments, this method proved to be rather fast for small-scale problems, but more inaccurate than other alternatives. This makes it a good choice for initialization.
	\item Given the initialization $\vect{u}$, solve the convex problem
	$\min_{\vect{v} \geq \vect{0}}\, \norm{\mat{D}_{\vect{u}} \left( \mat{A} - \vect{u} \transpose{\vect{v}} \right) }{1}$,
	using the ADMM technique in \cite[App.~A]{Tepper2014consensus} ($\mat{D}_{\vect{x}}$ is a diagonal matrix with the indicator vector $\vect{1}_{ [\vect{x} > 0] }$ in its diagonal).
	\item Given the new value of  $\vect{v}$, solve the convex problem
	$\min_{\vect{u} \geq \vect{0}}\, \norm{ \left( \mat{A} - \vect{u} \transpose{\vect{v}} \right) \mat{D}_{\vect{v}} }{1}$,
	using the same technique as before.
	\label{procedure:l1-nmf_speedup}
\end{compactenum}
The method in \cite{Kong2011} is not particularly well suited for large-scale problems, as it deals with \emph{dense} weighting matrices with the size of $\mat{A}$. As this size increases, handling these matrices becomes computationally prohibitive. \cref{sec:arse} is devoted to present a new technique to accelerate the above algorithm, rendering it capable of handling large-scale instances and producing a novel efficient L1-NMF solver.

\subsection{Dealing with multiple biclusters}

A challenge with biclustering (as with classical clustering) is that the number of biclusters is not an easy parameter to set or to estimate in advance.
Following a standard approach in the literature~\cite{papalexakis2013,witten2009ssvd} we sieve the information in a sequential way.
Generically, we iterate two steps until some stopping criterion is met: (1) find one bicluster $(\vect{u}, \vect{v})$, and (2) subtract the information encoded by $(\vect{u}, \vect{v})$ from $\mat{A}$.

Algorithm~\ref{algo:biclustering} summarizes the proposed biclustering approach. In \cref{algo_line:biclustering_remove_columns} we set to zero the columns corresponding to the active set of $\vect{v}_t$. This enforces disjoint active sets between the successive $\vect{v}_t$, and hence orthogonality. This also ensures that non-negativity is maintained throughout the iterations. The proposed algorithm is very efficient, simple to code, and demonstrated to work well in the experimental results that we will present later.

The iterations should stop (1) when $\mat{A}$ is empty (\cref{algo_line:biclustering_break_A}), or (2) when $\mat{A}$ only contains noise (no structured patterns). We deal with the second case in \cref{algo_line:biclustering_break_v}, considering that any bicluster formed by a single model instance is spurious.

\begin{algorithm2e}[t]
	%    \DontPrintSemicolon
	\SetKwInOut{Input}{input}\SetKwInOut{Output}{output}
	
	\begin{small}
		\Input{matrix $\mat{A} \in \Real^{m \times n}$}
		\Output{set of biclusters $\{ (\vect{u}_t, \vect{v}_t) \}_{t=1}^{T}$}
		
		$\mat{A}_{1} \gets \mat{A}$\; \label[algoline]{algo_line:biclustering_core_begin}
		\ForEach{$t \in \{ 1 \dots n \}$}{
			\If{$\mat{A}_{t} = \mat{0}$}{
				\label[algoline]{algo_line:biclustering_break_A}
				break\;
			}
			
			Find one bicluster $(\vect{u}_t, \vect{v}_t)$ of $\mat{A}_t$, where $\vect{u}_t \in \Real_{+}^m$ and $\vect{v}_t \in \Real_{+}^n$\;
			
			\nosemic
			$
			(\forall i, j)\
			(\mat{A}_{t+1})_{ij} \gets 
			\begin{cases}
			0 & \text{if } (\vect{v}_t)_{j} > 0 \text{,} \\
			(\mat{A}_{t})_{ij} & \text{otherwise;}
			\end{cases}
			$\;
			\label[algoline]{algo_line:biclustering_remove_columns}
			
%			\If{models are not allowed to share elements}{
%				$
%				(\forall i, j)\ 
%				(\mat{A}_{t+1})_{ij} \gets 
%				\begin{cases}
%				0 & \text{if } (\vect{u}_t)_{i} > 0 \text{,} \\
%				(\mat{A}_{t})_{ij} & \text{otherwise;}
%				\end{cases}
%				$\;
%				\label[algoline]{algo_line:biclustering_remove_rows}				
%			}
			\dosemic
			\If{$\norm{\vect{v}_t}{0} \leq 1$}{
				\label[algoline]{algo_line:biclustering_break_v}
				break\;
			}			
		}		
		\label[algoline]{algo_line:biclustering_core_end}
		
		\If(\tcp*[h]{if early stop, $t-1$ is the last good bicluster}){$t < n$}{
			$T \gets t - 1$\;
		}		
		Prune $\{ (\vect{u}_t, \vect{v}_t) \}_{t=1}^{T}$ using a model selection strategy\;
		\label[algoline]{algo_line:biclustering_prune}
	\end{small}
	
	\caption{Sequential rank-one biclustering algorithm.}
	\label{algo:biclustering}
\end{algorithm2e}

\subsection{Minimum description length as a stopping criterion}

The core of \cref{algo:biclustering} (\crefrange{algo_line:biclustering_core_begin}{algo_line:biclustering_core_end}) does not provide a \emph{reliable} mean to determine the number of biclusters to be extracted from the preference matrix. Indeed, \cref{algo_line:biclustering_break_A,algo_line:biclustering_break_v} only provide a \emph{rough} stopping criterion when $\mat{A}_{t+1}$ is empty or when the bicluster is not the product of a consensus between at least two models. This criterion will output more models than needed, and this section is devoted to prune the collection $\{ (\vect{u}_t, \vect{v}_t) \}_{t=1}^{T}$.

For each bicluster, we are only interested in the support of its vectors $\vect{u}_t$ and $\vect{v}_t$ (which are sparse by design). We can then binarize them to avoid being mislead by small numerical inaccuracies that might have occurred during the optimization procedure, i.e., $\ubar{\vect{u}}_t = \lfloor \vect{u}_t \rfloor_{\gamma} $ and $\ubar{\vect{v}}_t = \lfloor \vect{v}_t \rfloor_{\gamma}$, where
\begin{equation}
	(\lfloor \vect{x} \rfloor_{\gamma})_i = 
	\begin{cases}
		1 & \text{if } (\vect{x})_i > \gamma \cdot \norm{\vect{x}}{\infty}; \\
		0 & \text{otherwise.}
	\end{cases}
	\label{eq:truncation}
\end{equation}
In practice, we set $\gamma = 10^{-4}$ once and for all the tests in this paper.

We now pose the following model selection problem:
given the biclusters $\{ (\ubar{\vect{u}}_t, \ubar{\vect{v}}_t) \}_{t=1}^{T}$ and the remainders $\{ \mat{A}_{t+1} \}_{t=1}^{T}$ (which are binary by definition), find the value $K$ in $[1, T]$ such that the preference matrix $\mat{A}$ is \emph{optimally} described by
$\mat{A} \approx \sum_{t=1}^{K} \ubar{\vect{u}}_t \transpose{\ubar{\vect{v}}_t} + \mat{A}_{K+1}$.
This can be considered as an hypothetical compression problem where the task is to encode $\mat{A}$ using these elements. The model selection procedure then keeps the value $K$ that minimizes the combined codelenghts $L$ (in bits) of its components,
\begin{equation}
\argmin_{1\leq K < T} \sum_{t=1}^{K} L (\ubar{\vect{u}}_t) + \sum_{t=1}^{K}  L (\ubar{\vect{v}}_t) + L(\mat{A}_{K+1}).
\end{equation}
Under the usual assumption that $\ubar{\vect{u}}_t$, $\ubar{\vect{v}}_t$, and $\mat{A}_{K+1}$ are individually decorrelated, we can describe each one as a (one dimensional) i.i.d. Bernoulli sequence of values. For a $p$-dimensional vector $\vect{p}$, this can be efficiently described using an enumerative code~\cite{cover1973},
\begin{align}
	L(\vect{p}) = \log_2 \binom{p}{\norm{\vect{p}}{0}} + \log_2 p ,
\end{align}
where $\norm{\cdotp}{0}$ denotes number of non-zero elements of the argument and $\binom{a}{b}$ is the binomial coefficient. With a slight abuse of notation, for a matrix $\mat{M}$ we write $L(\mat{M}) = L(\operatorname{vec}(\mat{M}))$, where $\operatorname{vec}(\cdotp)$ is a vectorization operator.

\subsection{Statistical validation for pre-processing}

The standard random sampling approach (\cref{algo:random_sample}) to multiple model estimation generates many good model instances (composed of inliers), but also generates many bad models (composed mostly of outliers). In general, the number of bad models exceeds by far the number of good ones. It is not worth devoting computational effort to the analysis of these columns of $\mat{A}$.
Any pattern-discovery technique, such as the biclustering approach presented in the previous section, would benefit from having a simple, efficient, and statistically meaningful method for discarding bad models.
These models will typically contain only a handful of objects. The question is how do we determine the minimum size for a good consensus set?
This important computational contribution, based on the a contrario testing mechanism presented in depth in~\cite{desolneux08}, is addressed next.

Let us assume that we have a set $\set{X}$ of random elements.
Let $\mu(\theta)$ be a model with an associated consensus set $\set{C}_{\mu} (\set{X}, \theta, \delta)$, built with tolerance $\delta$ (\cref{eq:consensusSet}, \cpageref{eq:consensusSet}).
We will assume, under the background model, that all objects are i.i.d. and that the error in \cref{eq:pointModelDistance} \emph{locally} follows an uniform distribution; this type of simple approximations has proven successful for outlier rejection~\cite{desolneux08}.
Let $\kappa$ be a locality parameter. If there are $\left| \set{C}_{\mu} \left(\set{X}, \theta, \delta \right)\right|$ elements at a distance $\delta$ from $\mu(\theta)$, we expect to have in average $\kappa$ times more elements at a distance $\kappa\delta$.
We are interested in computing the probability that $\set{C}_{\mu} (\set{X}, \theta, \delta)$ has at least $m_{\delta}$ elements given that $\set{C}_{\mu} (\set{X}, \theta, \kappa\delta)$ contains $m_{\kappa\delta}$ elements.
The probability of such an event is $\bintail{m_{\kappa\delta} - b}{m_{\delta} - b}{p}$, where $\mathcal{B}$ is the binomial tail and $p = \delta / \kappa\delta = \kappa^{-1}$ is the probability that a random object belongs to the consensus set $\set{C}_{\mu} (\set{X}, \theta, \delta)$ given that it belongs to $\set{C}_{\mu} (\set{X}, \theta, \kappa\delta)$.
Recall that $b$ is the minimum number of elements necessary to uniquely characterize a given parametric model. As such, $b$ elements in $\set{C}_{\mu} (\set{X}, \theta, \kappa\delta)$ are necessarily non-random, since they were used to estimate $\mu(\theta)$. This is the reason behind subtracting $b$ elements from $m_{\kappa\delta}$ and $m_{\delta}$.

\begin{definition}
	Let $\mu(\theta)$ be a model instance, $\set{C}_{\mu} (\set{X}, \theta, \delta)$ be its associated consensus set, obtained with precision parameter $\delta$, and $\kappa > 1$ be a locality parameter.
	We define the number of false Alarms (NFA) of model instance $\mu(\theta)$ as
	\begin{equation}
		\operatorname{NFA}_{\mu, \delta, \kappa} (\set{X}, \theta) = N_\text{tests} \cdot \bintail{\left| \set{C}_{\mu} \left(\set{X}, \theta, \kappa\delta \right)\right| - b}{\left| \set{C}_{\mu} \left(\set{X}, \theta, \delta \right)\right| - b}{\kappa^{-1}},
	\end{equation}
	where $N_\text{tests} = \binom{m}{b}$ represents the total number of possible models.
	We say that the model $\mu(\theta)$ is said to be \meps-meaningful if
	\begin{equation}
		\operatorname{NFA}_{\mu, \delta, \kappa} (\set{X}, \theta) < \eps.
	\end{equation}
	\label[definition]{def:meaningful}
\end{definition}
It is easy to prove, by the linearity of expectation, that the expected number of \meps-meaningful models in a finite set of random models is smaller than \meps.
Alternatively, $N_\text{tests}$ can be empirically set by analyzing a training dataset~\cite{burrus2009pr}, providing a tighter bound for the expectation.

\Cref{def:meaningful} provides a formal probabilistic method for testing if a model is likely to happen at random or not. From a statistical viewpoint, the method goes back to multiple hypothesis testing.
Following an a contrario reasoning~\cite{desolneux08}, we decide whether the event of interest has occurred if it has a very low probability of occurring by chance in the above defined random (background) model. In other words, a model instance $\mu(\theta)$ is \meps-meaningful if $|\set{C}_{\mu} (\set{X}, \theta, \delta)|$ is sufficiently large to have $\operatorname{NFA}_{\mu, \delta, \kappa} (\set{X}, \theta) < \eps$. We only keep columns of $\mat{A}$ corresponding \meps-meaningful model instances. We set $\eps=1$ for all the experiments.

As a result of this statistical validation procedure, the preference matrix $\mat{A}$ is considerably shrunk (the actual size reduction will depend on the inliers-outliers ratio and the number of model instances in the dataset). This shrunk preference matrix is fed to the biclustering algorithm (Algorithm~\ref{algo:biclustering}), gaining in stability of the results as well as in speed.

\begin{note}
	There are techniques in the literature \cite[e.g.,][]{chin2012}, which follow a different route: instead of sampling first and prune later, they try to dynamically sample \emph{good} models from the start. These techniques have proven successful in reducing the number of required samples but operate at a much slower rate per sample. Their dynamic nature makes parallelization more difficult and limited. In our view, a detailed comparison between these paradigms is needed, with special consideration given to the use of parallelization (e.g., GPUs).
\end{note}

\subsection{Statistical validation for post-processing}

Once the biclustering algorithm has returned a collection of inliers-model pairs, we need to verify that these models are statistically meaningful from a geometric point of view. For this, we use the test in \cref{def:meaningful} once again.

\begin{figure}
	\centerline{
		\hfill
		\includegraphics[width=0.3\textwidth]{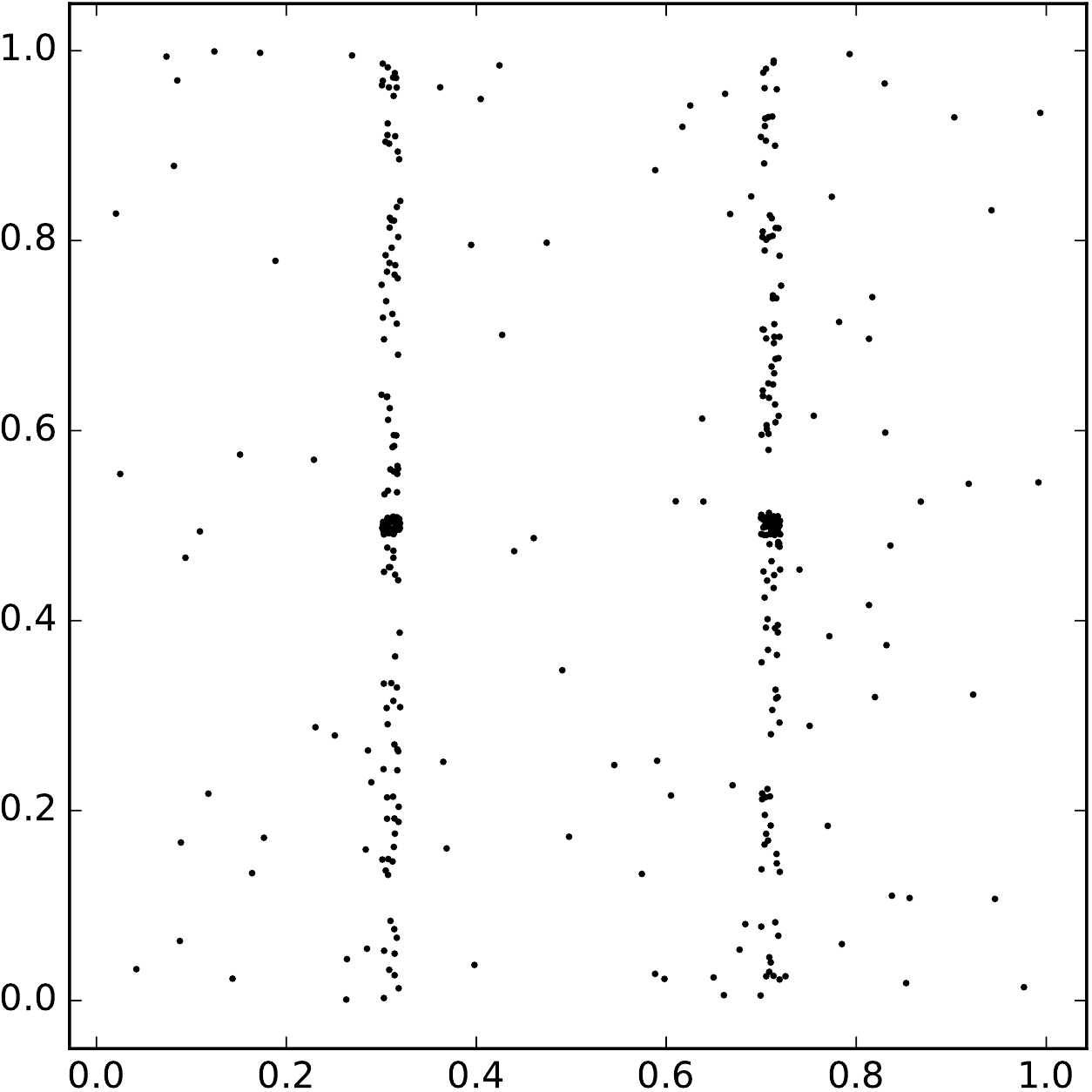}
		\hspace{.5cm}
		\includegraphics[width=0.3\textwidth]{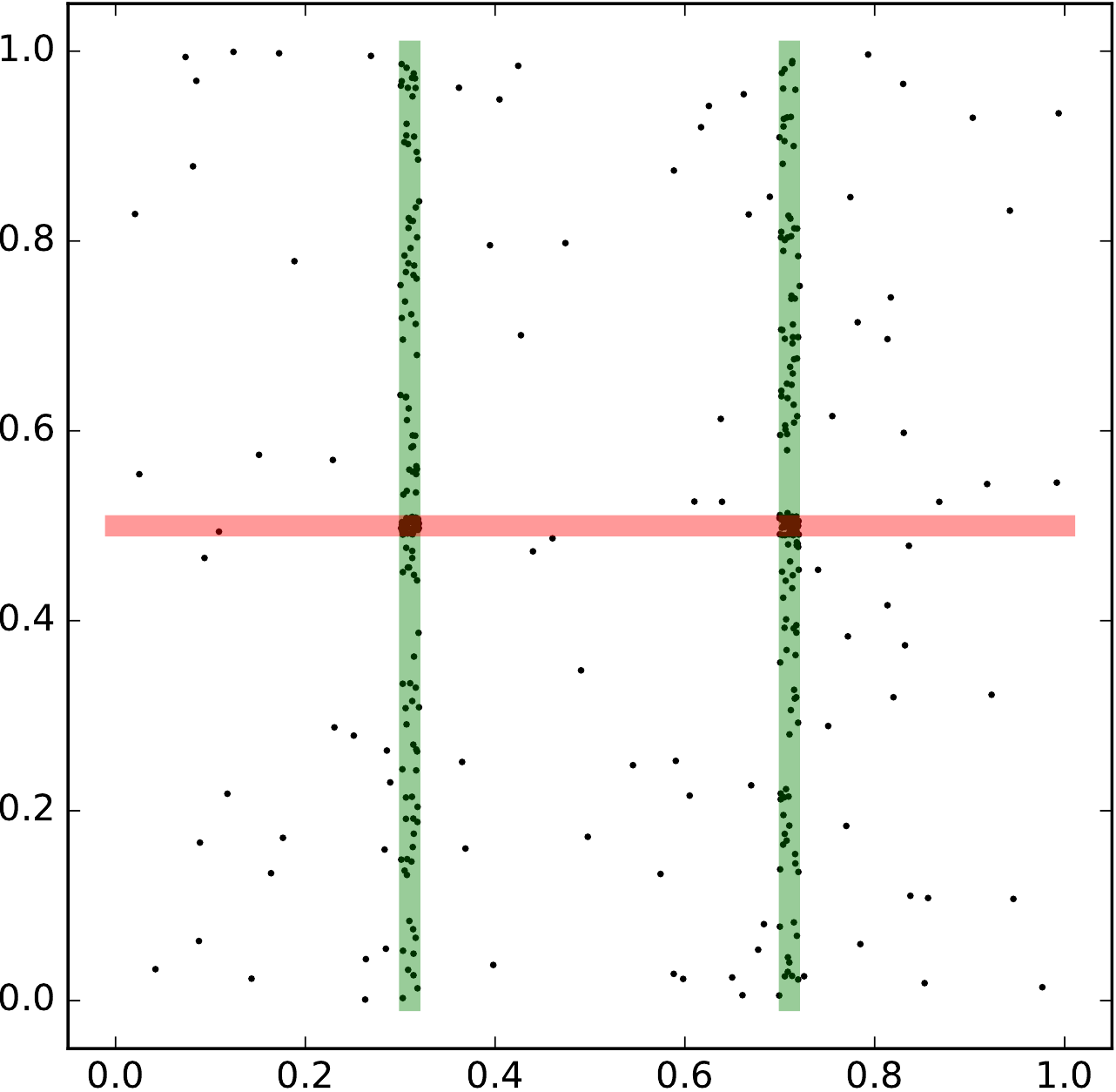}
		\hfill
	}
	
	\caption{The necessity of an exclusion principle. On the left, an arrangement of points built from uniformly sampling 100 points in $[0, 1]^2$, 100 points in $[0.29, 0.31] \times [0, 1]$, 100 points in $[0.69, 0.71] \times [0, 1]$, 50 in $[0.29, 0.31] \times [0.49, 0.51]$, and 50 points  in $[0.69, 0.71] \times [0.49, 0.51]$. On the right, three plausible detections from a hypothetical MPME algorithm. We would like to keep the green lines and discard the red one, which is merely an artifact stemming from the definition of a consensus set (\cref{eq:consensusSet}). Note that, after removing the intersection with the green line, there is nothing that distinguishes the red band from the background.}
	\label{fig:exclusion_principle}
\end{figure}

Additionally, models are allowed to overlap and can thus share elements, which makes situations like the one described in \cref{fig:exclusion_principle} commonplace.
In short, a good model should separate itself from the background noise \emph{regardless} of its intersection with other models.
The exclusion principle described in~\cite[Chapter~6]{desolneux08} performs this check. It states that, given two groups obtained by the same detector, no point $\vect{x}$ is allowed to belong to both groups.
In our case, since we are explicitly modeling overlaps between groups, we use a milder version of this principle: we simply ask that any point can only contribute to the NFA of at most one group. This is explained in \cref{algo:exclusion_principle}. Notice that the order in which the inliers-models pairs are explored affects the result; we sort the biclusters according to their total size $\norm{\ubar{\vect{u}}_t}{0} \cdot \norm{\ubar{\vect{v}}_t}{0}$ (\cref{eq:truncation}).

\begin{algorithm2e}[t]
	\SetKwInOut{Input}{input}
	\SetKwInOut{Output}{output}
	
	\begin{small}
	\Input{collection of inliers-model pairs $\{ (\set{C}_t, \theta_t) \}_{t=1}^{T}$.}
	\Output{pruned collection of inliers-model pairs $\{ (\set{C}_t, \theta_t) \}_{t=1}^{T}$.}
	
	$\set{K} \gets \emptyset$\;
	$\set{S} \gets \set{X}$\;
	\ForEach{$(\set{C}_t, \theta_t)$}{
		\If{$\operatorname{NFA}_{\mu} (\set{S}, \theta, \delta)$ is meaningful}{
			$\set{K} \gets \set{K} \cup \{ (\set{C}_t, \theta_t) \}$\;
			$\set{S} \gets \set{X} \setminus \set{C}_t$\;			
		}
	}
	\end{small}

	\caption{Exclusion principle}
	\label{algo:exclusion_principle}
\end{algorithm2e}

Optionally and if required by the application, as the last step in our post-processing chain, we can force the models to have an empty intersection. There are many alternative ways to address this assignment problem. In this work, we simply assign elements in the intersection of several models to the closest model in distance (see \cref{eq:pointModelDistance}).

\section{Accelerated Random Sample Ensemble}
\label{sec:arse}

The problem of multiple parametric model estimation does not escape the current trend of growth in datasets size, which exposes the need of fast techniques to cope with these massive datasets. Thus, after having described the proposed Random Sample Ensemble algorithmic pipeline in \cref{sec:rse}, we now turn our attention to its acceleration. There are two computational bottlenecks in our approach.

The first one is shared by virtually all the algorithms in the field: running \cref{algo:random_sample} (\cpageref{algo:random_sample}). Fortunately, all the random samples can be computed in parallel, reducing the problem to clever software engineering. Alternatively, there are recent techniques to reduce the number of needed samples \cite{chin2012}, at the expense of a less parallelizable algorithm.

Second, the main component of the proposed technique is the biclustering algorithm, and this section is thus devoted to describe how to efficiently solve this optimization problem. Let us remind the reader that we are seeking to solve \cref{eq:l1-nmf} (\cpageref{eq:l1-nmf}), a challenge with applications beyond the problem at hand.
Before moving forward with the exposition, we need to lay the ground by providing a few definitions and key concepts.

\paragraph{The fast Cauchy transform.}
The fast Johnson-Lindenstrauss transform~\cite{Ailon2006,Ailon2009} provides a way to build a low dimensional embedding in the $\ell_2$ case and has been widely used in many practical settings. Its $\ell_1$-based analog is the fast Cauchy transform (FCT) \cite{Clarkson2016}, which defines an $h \times m$ embedding matrix $\mat{\Pi}$ ($h \ll m$) as
\begin{equation}
	\mat{\Pi} = 4 \mat{B} \mat{C} \widetilde{\mat{H}} .
	\label{eq:fct}
\end{equation}
The matrices $\mat{B}$, $\mat{C}$, and $\widetilde{\mat{H}}$ are built such that
\begin{itemize}
	\item each column of $\mat{B} \in \Real^{h \times 2m}$ is chosen at random from the $h$ standard basis vectors for $\Real^{h}$;
	\item $\mat{C} \in \Real^{2m \times 2m}$ is a diagonal matrix with entries sampled from a Cauchy distribution; and
	\item $\widetilde{\mat{H}} \in \Real^{2m \times m}$ is a block-diagonal matrix comprised of $m/s$ equal blocks along the diagonal (we set $s=h^6$ and we assume it to be a power of two and $m/s$ is an integer), i.e.,
	\begin{align}
	\widetilde{\mat{H}} &\defeq
	%        \left[
	\begin{array}{@{}|cccc|@{}}
	\hline
	\multicolumn{1}{@{}|c|@{}}{
		\begin{matrix}
		\mat{G}_s
		\end{matrix}
	} &&& \\
	\cline{1-2}
	&
	\multicolumn{1}{@{}|c|@{}}{
		\mat{G}_s
	}  && \\
	\cline{2-2}
	&& \ddots & \\
	\cline{4-4}
	&&&
	\multicolumn{1}{@{}|c|@{}}{
		\mat{G}_s
	}\\
	\hline
	\end{array}
	&
	\text{where}
	&&        
	\mat{G}_s &\defeq
	\begin{bmatrix}
	s^{-1/2} \cdot \mat{H}_s \\
	\mat{I}_s
	\end{bmatrix} ,
	\end{align}
	$\mat{I}_s$ is the $s \times s$ identity matrix, and $\mat{H}_s$ is the $s \times s$ Hadamard matrix, defined recursively as
	\begin{align}
	\mat{H}_2 &=
	\begin{bmatrix}
	+1 & +1 \\
	+1 & -1
	\end{bmatrix},
	&
	\mat{H}_s &=
	\begin{bmatrix}
	\mat{H}_{s/2} & \mat{H}_{s/2} \\
	\mat{H}_{s/2} & -\mat{H}_{s/2}
	\end{bmatrix}.
	\end{align}
\end{itemize}
The following theorem provides some guarantees about the FCT.

\begin{theorem}[\cite{Clarkson2016}]
	Let $\mat{A} \in \Real^{m \times n}$ be a matrix of rank $r$ ($r \ll n$).
	There is a distribution (given by the above FCT construction) over matrices $\mat{\Pi} \in \Real^{h \times m}$ with $h = O(r \log r + r \log \tfrac{1}{\eta})$ such that, for all $\vect{x} \in \Real^{n}$, the inequalities
	\begin{equation}
	\norm{\mat{A}\vect{x}}{1} \leq \norm{\mat{\Pi} \mat{A}\vect{x}}{1} \leq \tau \norm{\mat{A}\vect{x}}{1}
	\label{eq:l1_projection_bounds}
	\end{equation}
	hold with probability $1- \eta$, where
	\begin{equation}
	\tau = O \left( \frac{r \sqrt{s}}{\eta} \log (hr) \right) .
	\end{equation}
\end{theorem}
Considering $\eta$ as a (small) constant and recalling that $s = h^6$, we have $h = O(r \log r)$ and finally $\kappa = O \left( r^4 \log^4 r \right)$. However, if we set $s \ll h^6$, the performance in practice does not seem to be negatively affected, even if this setting does not follow the above theorem \cite[Section~6.1]{Clarkson2016}. In our experiments, we use $s = h$.

\subsection{Fast $\ell_1$ regression}

The FCT can be used as a building block for the construction of fast solvers for $\ell_1$ regression problems. We describe this first before presenting the L1-NMF proposed algorithm.

\begin{definition}[\cite{Sohler2011}]
	Let $\mat{A} \in \Real^{m \times n}$ be a matrix of rank $r$.
	A basis $\mat{B} \in \Real^{n \times r}$ for the range of $\mat{A}$ is $(\alpha, \beta)$-conditioned if $\norm{\mat{B}}{1} \leq \alpha$ and $(\forall \vect{x} \in \Real^{r})\ \norm{\vect{x}}{\infty} \leq \norm{\mat{B} \vect{x}}{1}$. We say that $\mat{B}$ is well conditioned if $\alpha$ and $\beta$ are low degree polynomials in $r$, independent of $m$ and $n$.
	\label[definition]{def:well_conditioned_basis}
\end{definition}

\begin{definition}[\cite{Sohler2011}]
	Given a well-conditioned basis $\mat{B}$ for the range of $\mat{A} \in \Real^{m \times n}$, we define the $\ell_1$ leverage scores of $\mat{A}$ as the $m$-dimensional vector $\vect{\lambda}$, with elements $(\vect{\lambda})_i = \norm{(\mat{B})_{:i}}{1}$.
	\label[definition]{def:leverage_scores}
\end{definition}

\noindent
The leverage scores of $\mat{A}$ can be found with the following procedure \cite{Clarkson2016,Sohler2011}:
\begin{compactenum}
	\item build an FCT matrix $\mat{\Pi} \in \Real^{r \times m}$;
	\item find a matrix $\mat{R}$ such that $\mat{\Pi} \mat{A} \mat{R}$ is orthonormal;
	\item using \cref{def:leverage_scores}, compute the leverage scores $\vect{\lambda}$ of $\mat{B} = \mat{A} \mat{R}^\dagger$ ($\mat{R}^\dagger$ denotes the pseudoinverse of $\mat{R}$);		
\end{compactenum}
The leverage scores are used in \cite{Clarkson2016,Sohler2011} to speed up the algorithmic solution of
$\vect{x}^* = \argmin_{\vect{x}} \norm{\mat{A} \vect{x} - \vect{b}}{1}$.
For this, the authors build a diagonal matrix $\mat{E}$, i.e., a compression matrix, by sampling its diagonal entries independently according to a set of probabilities $p_i$ that are proportional to the $\ell_1$ leverage scores of $\mat{A}$. Then, given
$\hat{\vect{x}} = \argmin_{\vect{x}} \norm{\mat{E}(\mat{A} \hat{\vect{x}} - \vect{b})}{1}$
we have that
$\norm{\mat{A} \hat{\vect{x}} - \vect{b}}{1} \leq (1 + \epsilon) \norm{\mat{A} \vect{x}^* - \vect{b}}{1}$ for some small $\epsilon$. Key to the speedup is the size reduction of the regression problem, since only a few rows of $\mat{A}$ are kept in $\mat{E} \mat{A}$ and considered to find $\hat{\vect{x}}$.

\subsection{Fast $\ell_1$ NMF}
 
We now have all the elements that we need to build a fast algorithm for solving \cref{eq:l1-nmf}. Considering that we are dealing with a biconvex problem, it is a standard practice to find a solution to it by alternating two $\ell_1$ least squares problems,
\begin{subequations}
\begin{align}
	\hat{\vect{u}} &= 
	\arg\min_{\vect{u}} \norm{\mat{A} - \vect{u} \transpose{\hat{\vect{v}}}}{1}
	\quad \text{s.t.} \quad
	\vect{u} \geq \vect{0},
	\\
	\hat{\vect{v}} &= 
	\arg\min_{\vect{v}} \norm{\mat{A} - \hat{\vect{u}} \transpose{\vect{v}}}{1}
	\quad \text{s.t.} \quad
	\vect{v} \geq \vect{0}.
\end{align}
\end{subequations}
Each sub-problem can be sped up by respectively compressing the rows and the columns of $\mat{A}$. Similarly as in the regression case, row (resp.~column) compression is achieved through the computation of the leverage scores of $\mat{A}$ (resp.~$\transpose{\mat{A}}$). Let $\mat{C}$ and $\mat{R}$ be the matrices that perform row and column compression, respectively. We can then iterate the \emph{compressed} least squares problems
\begin{subequations}
\begin{align}
	\hat{\vect{u}} &= 
	\arg\min_{\vect{u}} \norm{\left( \mat{A} - \vect{u} \transpose{\hat{\vect{v}}} \right) \mat{C}}{1}
	\quad \text{s.t.} \quad
	\vect{u} \geq \vect{0},
	\\
	\hat{\vect{v}} &= 
	\arg\min_{\vect{v}} \norm{\mat{R} \left( \mat{A} - \hat{\vect{u}} \transpose{\vect{v}} \right)}{1}
	\quad \text{s.t.} \quad
	\vect{v} \geq \vect{0}.
\end{align}
\end{subequations}
This would already give a very efficient algorithm for solving the problem of interest.
In our experiments, we found that augmenting the procedure described in \cpageref{procedure:l1-nmf_speedup} with the above described compression techniques was faster and produced good results.
We thus obtain the following \emph{accelerated} procedure for solving \cref{eq:l1-nmf}:
\begin{compactenum}
	\item Given a column compression matrix $\mat{C}$ for $\mat{A}$, find an initialization for $\vect{u}$ by solving the \emph{compressed} problem
	\begin{equation}
		\hat{\vect{u}}, \hat{\vect{v}} = 
		\argmin_{\vect{u}, \tilde{\vect{v}}} \norm{\mat{A} \mat{C} - \vect{u} \transpose{\tilde{\vect{v}}} }{1}
		\quad \text{s.t.} \quad
		\vect{u} \geq \vect{0}, \tilde{\vect{v}} \geq \vect{0},
	\end{equation}
	with the iterative re-weighting scheme in \cite{Kong2011}. The compressed vector $\hat{\vect{v}}$ has no further use in our algorithm.
	\item Given the initialization $\hat{\vect{u}}$ and a row compression matrix $\mat{R}_{\hat{\vect{u}}}$ for $\mat{D}_{\hat{\vect{u}}} \mat{A}$ ($\mat{D}_{\vect{x}}$ is a diagonal matrix with the vector $\vect{1}_{ [\vect{x} > 0] }$ in its diagonal.), solve the \emph{compressed} convex problem
	$\min_{\vect{v} \geq \vect{0}}\, \norm{ \mat{R}_{\hat{\vect{u}}} \mat{D}_{\hat{\vect{u}}} \left( \mat{A} - \hat{\vect{u}} \transpose{\vect{v}} \right) }{1}$,
	using the ADMM technique in \cite[App.~A]{Tepper2014consensus}.
	\item Given the new value of  $\vect{v}$ and a column compression matrix $\mat{C}_{\vect{v}}$ for $\mat{A} \mat{D}_{\vect{v}}$, solve the \emph{compressed} convex problem
	$\min_{\vect{u} \geq \vect{0}}\, \norm{ \left( \mat{A} - \vect{u} \transpose{\vect{v}} \right) \mat{D}_{\vect{v}} \mat{C}_{\vect{v}} }{1}$,
	using the same technique as before.
	\label{procedure:acc_l1-nmf_speedup}
\end{compactenum}
All compression matrices are computed with the same compression level $r$, used to build the FCT matrix.
We also found in our experimental results that instead of using random sampling for building the row and column compression matrices, as in \cite{Clarkson2016}, good results were obtained by simply selecting the $r$ largest leverage scores.

\section{Experimental results}
\label{sec:results}

In the following, we refer to the proposed Random Sample Ensemble as RSE and to its compressed version, Accelerated Random Sample Ensemble, as ARSE. In our experiments, we compared the quantitative results of both methods, but only present qualitative results for ARSE. All the parameters for RSE and ARSE are exactly the same and in each figure/table we specify $\delta$ (\cref{eq:consensusSet}); unless specified, $\kappa = 3$ (\cref{def:meaningful}). The only exception is the compression rate in ARSE, which will be set to $h=32$ in all experiments, see \cref{eq:fct}.

\paragraph{Evaluation.}
To measure performance, we use the standard precision and recall.
In order to compute these measures, we compute an optimal assignment between the ground truth and the tested groups, using the Hungarian algorithm. Once the two sets of groups are appropriately matched, we can then compute the precision and the recall of the tested groups in a standard fashion.

If models are not allowed to share elements, it is not unusual in the literature to consider the outliers as an \emph{additional} ground truth group to recover; in this case, the \emph{misclassification error} is often reported.

In the general case where models overlap, we also use, as an additional measure, the generalized normalized mutual information (GNMI) \cite{lancichinetti2009gnmi}, which extends the normalized mutual information (also called symmetric uncertainty)~\cite[p. 310]{DataMining2011} to the case of groups with overlaps.

\begin{figure}[t]

	\begin{subfigure}{.2\linewidth}
		\includegraphics[height=.95\linewidth]{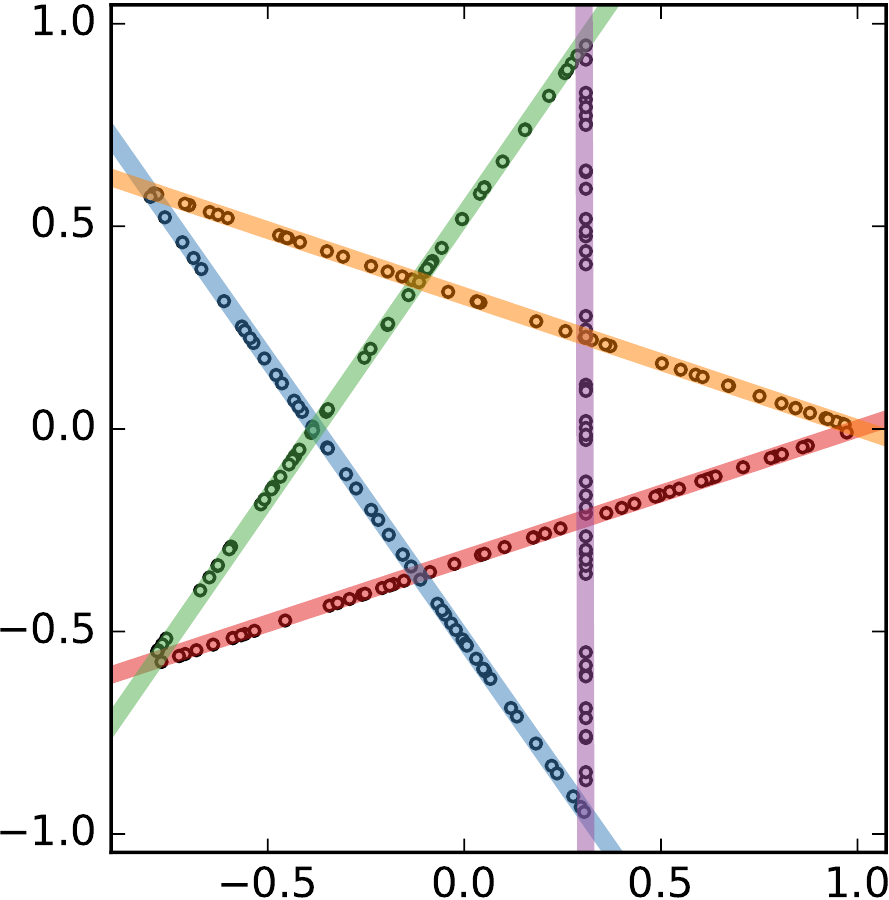}
		\caption{$\delta = 0.04$}
		\label{fig:Star5_S0_O5}
	\end{subfigure}%
	\hfill%
	\begin{subfigure}{.2\linewidth}
		\includegraphics[height=.95\linewidth]{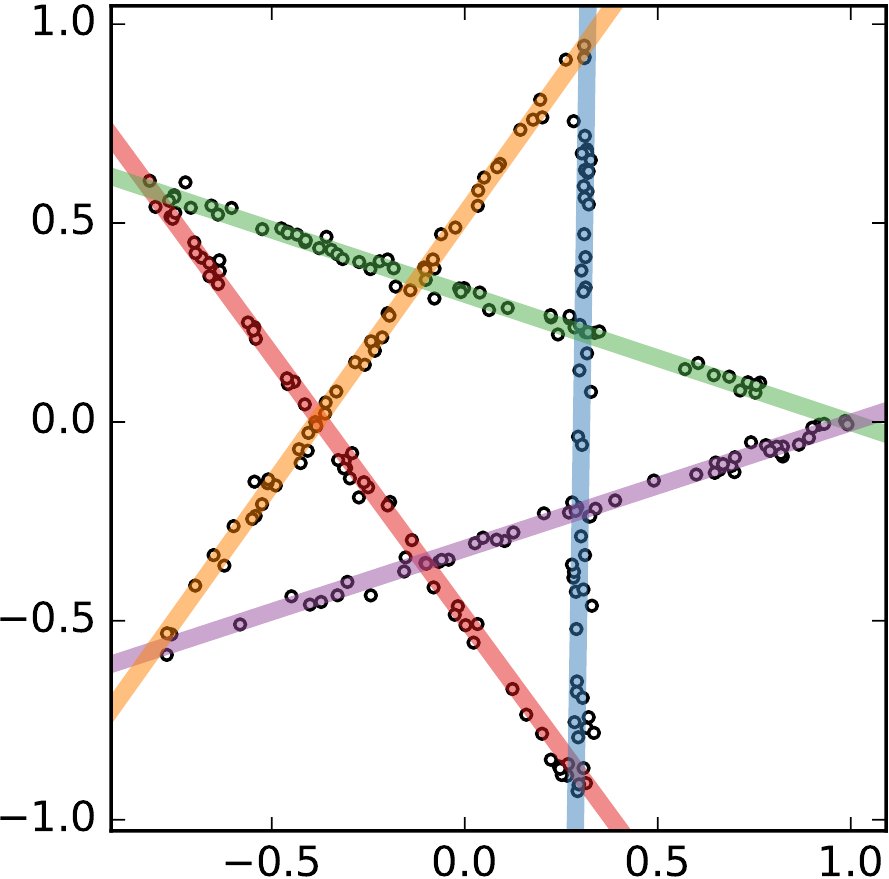}
		\caption{$\delta = 0.04$}
		\label{fig:Star5_S0015_O0}
	\end{subfigure}%		
	\hfill%
	\begin{subfigure}{.2\linewidth}
		\includegraphics[height=.95\linewidth]{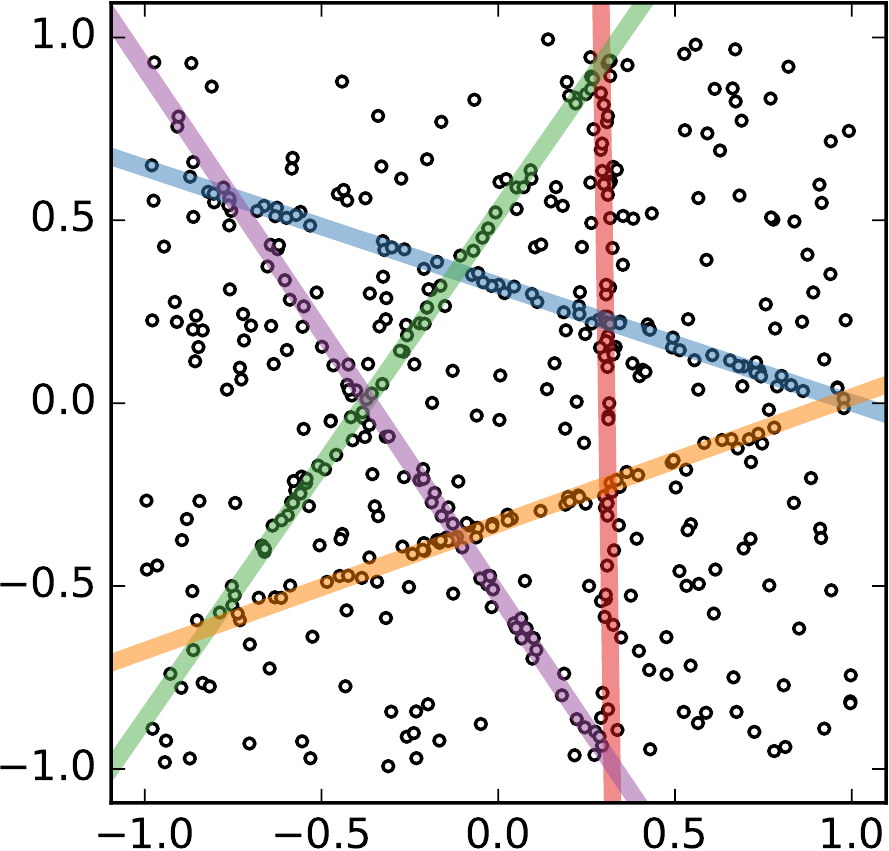}
		\caption{$\delta = 0.04$}
		\label{fig:Star5_S0015_O50}
	\end{subfigure}%
	\hfill%
	\begin{subfigure}{.2\linewidth}
		\includegraphics[height=.95\linewidth]{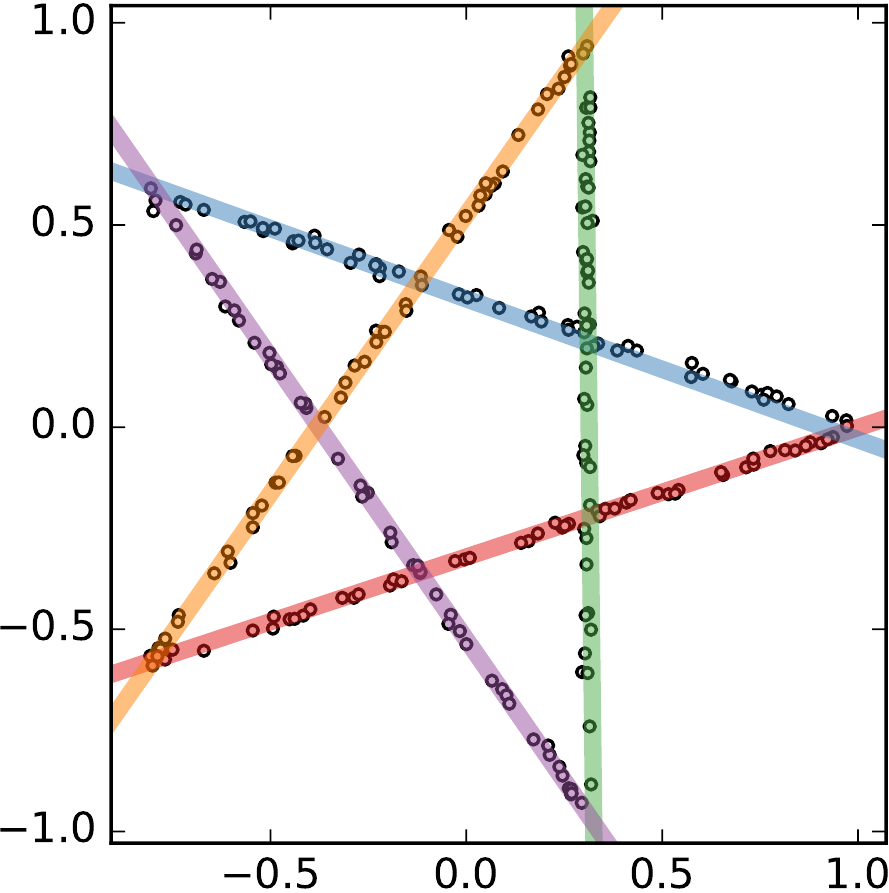}
		\caption{$\delta = 0.04$}
		\label{fig:Star5_S00075_O0}
	\end{subfigure}%
	\hfill%
	\begin{subfigure}{.2\linewidth}
		\includegraphics[width=.95\linewidth]{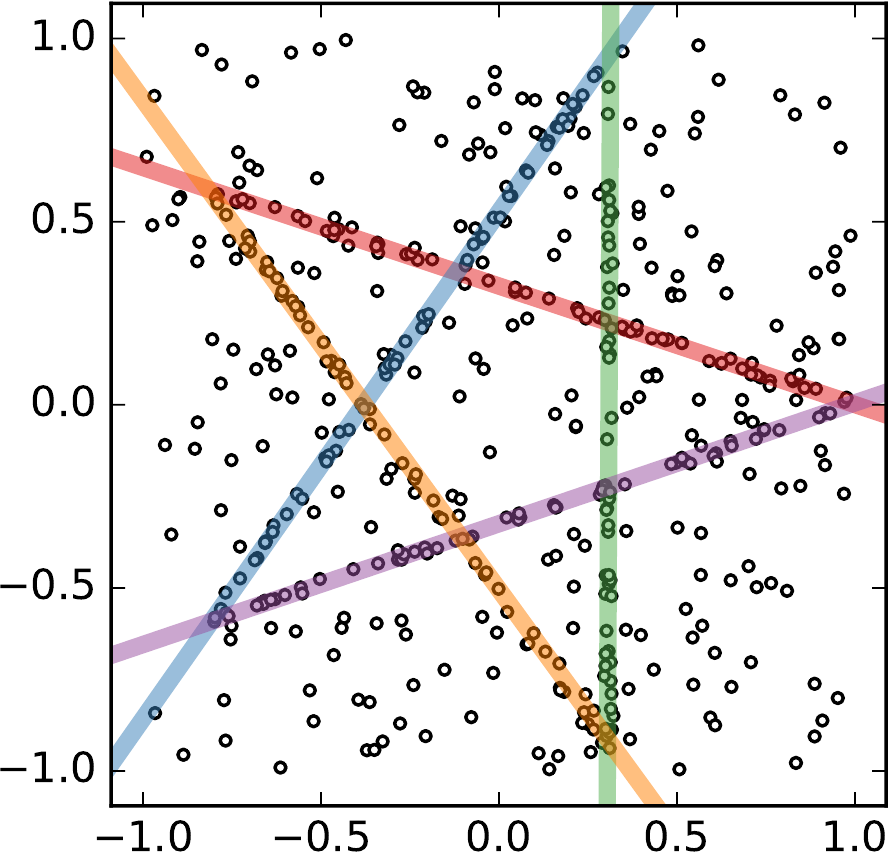}
		\caption{$\delta = 0.04$}
		\label{fig:Star5_S00075_O50}
	\end{subfigure}%
	\\
	\begin{subfigure}{.2\linewidth}
		\includegraphics[width=.95\linewidth]{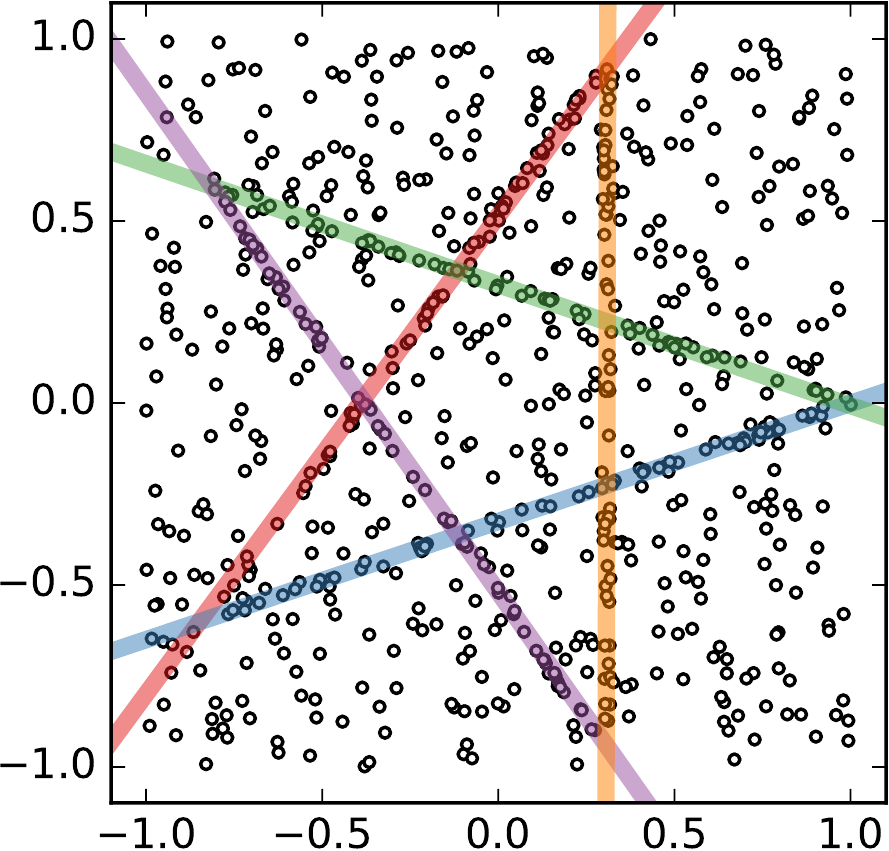}
		\caption{$\delta = 0.04$}
		\label{fig:Star5_S00075_O75}
	\end{subfigure}%
	\hfill%
	\begin{subfigure}{.2\linewidth}
		\includegraphics[width=.95\linewidth]{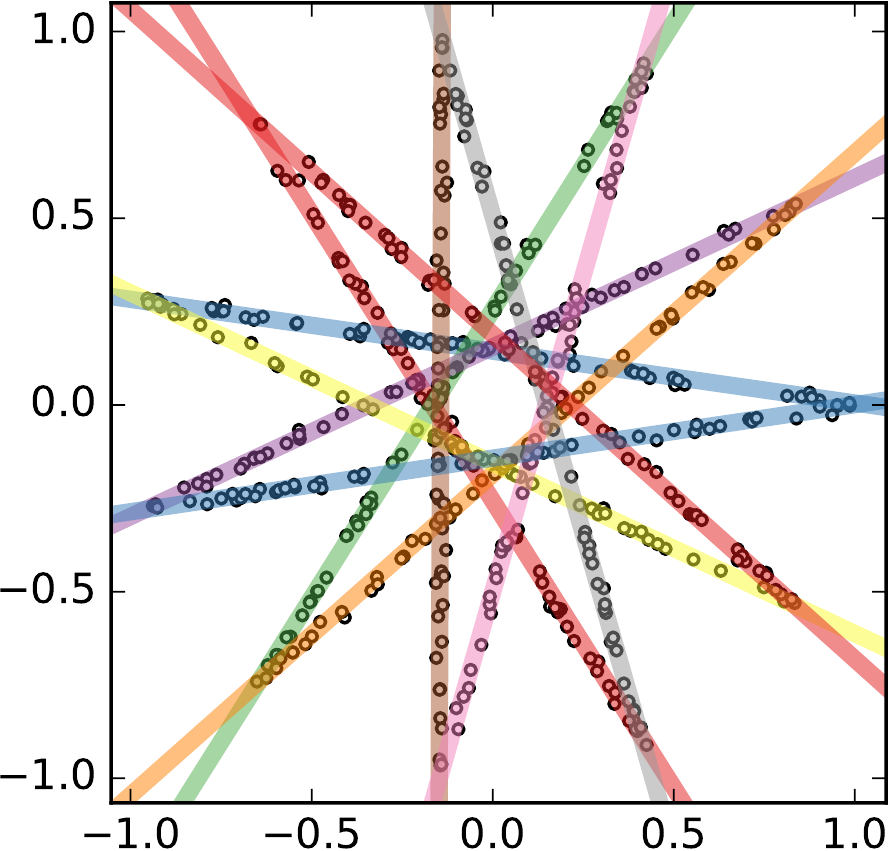}
		\caption{$\delta = 0.02$}
		\label{fig:Star11_S00075_O0}
	\end{subfigure}%
	\hfill%
	\begin{subfigure}{.2\linewidth}
		\includegraphics[width=.95\linewidth]{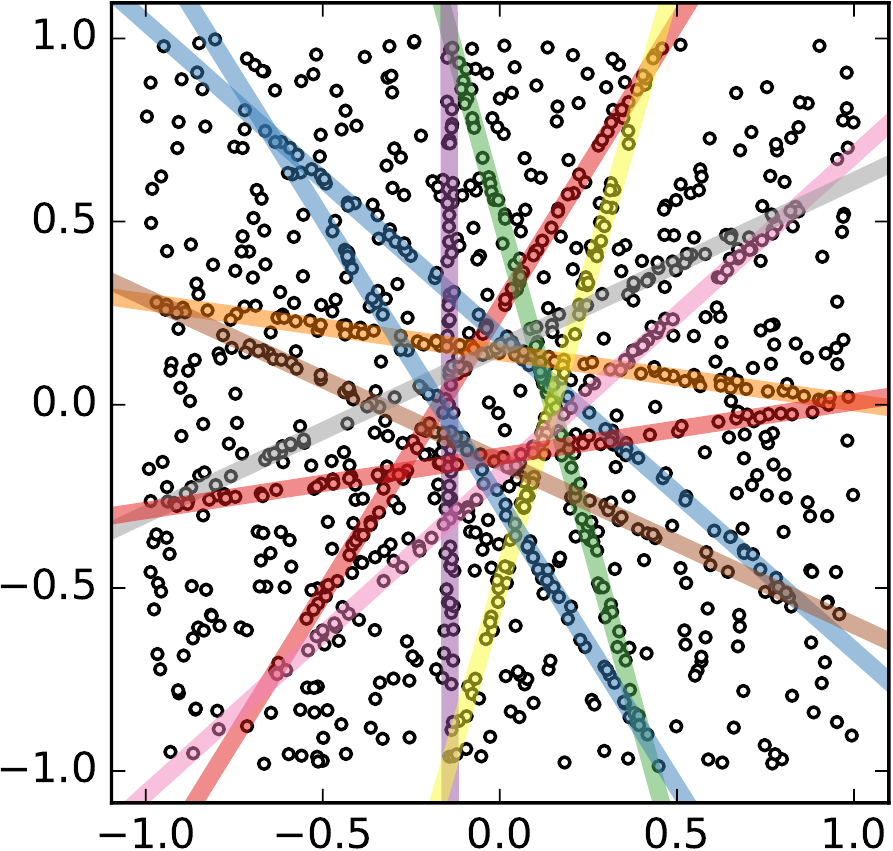}
		\caption{$\delta = 0.02$}
		\label{fig:Star11_S00075_O50}
	\end{subfigure}%
	\hfill%
	\begin{subfigure}{.2\linewidth}
		\includegraphics[width=.95\linewidth]{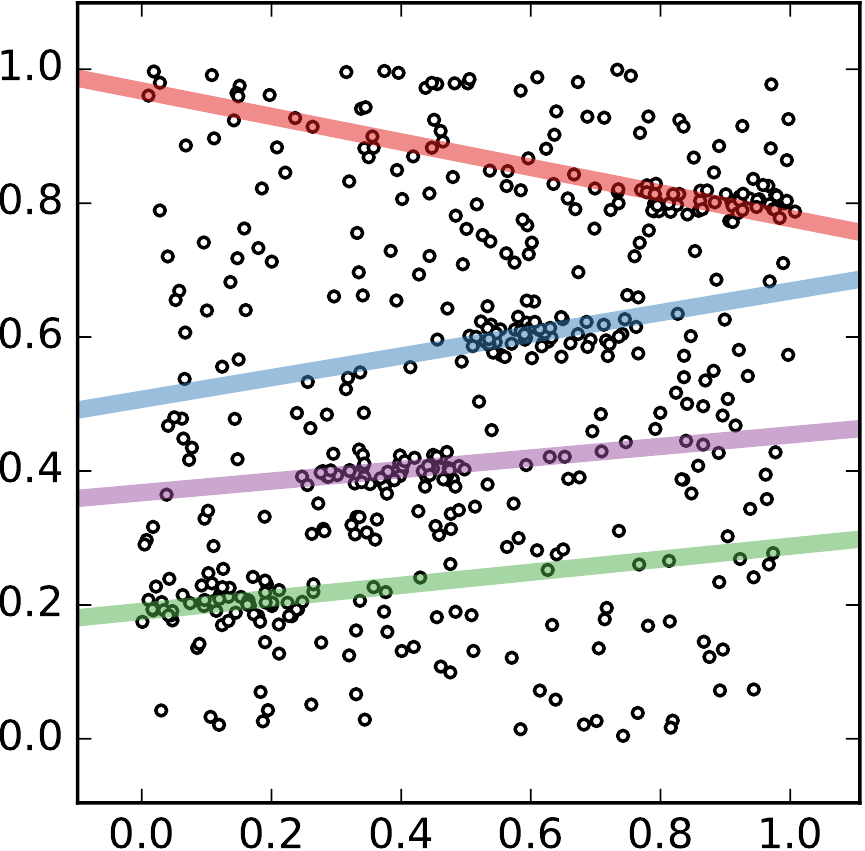}
		\caption{$\delta = 0.04$, $\kappa = 2$}
		\label{fig:Stairs_S0015_060}
	\end{subfigure}%
	\hfill%
	\begin{subfigure}{.2\linewidth}
		\includegraphics[width=.95\linewidth]{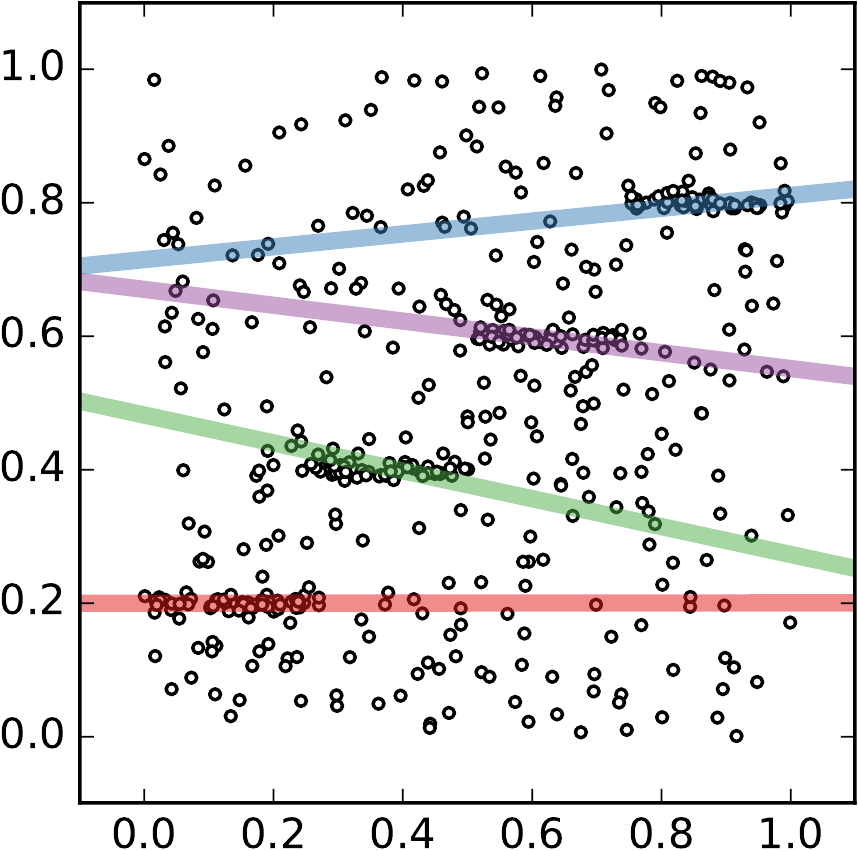}
		\caption{$\delta = 0.04$, $\kappa = 2$}
		\label{fig:Stairs4_S00075_O60}
	\end{subfigure}%
	
	\caption{ARSE results for 2D lines detection. We show with different colors the final models estimated from the extracted biclusters (some colors may repeat themselves).}
	\label{fig:test_2d}
\end{figure}

\begin{figure}[t]
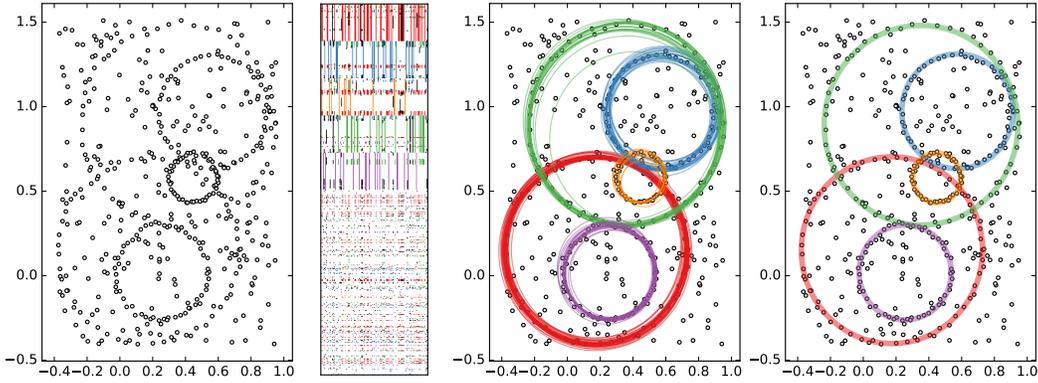

	
	\centering
	\begin{tabu} to .9\textwidth {@{\hspace{0cm}} X[c,m] @{\hspace{.1cm}} X[.5,c,m] @{\hspace{.1cm}} X[c,m] @{\hspace{.1cm}} X[c,m] @{\hspace{0cm}} }
		
		\gdef\fileprefix{test_2d_given_gt/Circles5_S00075_O50}
		\includegraphics[width=\linewidth]{{\fileprefix _data.pdf}} &
		\includegraphics[width=.76\linewidth]{\fileprefix _bic_comp_pref_mat.pdf} &
		\includegraphics[width=\linewidth]{\fileprefix _bic_comp_bundles.pdf} &			
		\includegraphics[width=\linewidth]{\fileprefix _bic_comp_final_models.pdf} \\
	\end{tabu}
	
	\caption{ARSE results for 2D circle detection on a synthetic dataset~\cite{toldo08}. From left to right, original dataset, preference matrix, models retained in each bicluster, and models estimated from each bicluster. Each model/bicluster is denoted with a different color (some colors may repeat themselves). Notice that in the biclustered preference matrix, the models actually overlap. For this example, we set $\delta = 0.02$.}
	\label{fig:test_2d_circles}
\end{figure}

\paragraph{2D lines and circles.}
We start our experimental evaluation with a few small synthetic datasets \cite{toldo08} where 2D points are arranged forming lines and circles.
The results are shown in \cref{fig:test_2d,fig:test_2d_circles}, where ARSE clearly detects the correct models in each case. In \cref{tab:test_2d} we compare the performance of RSE and ARSE in all these datasets. As expected, when the dataset (and the preference matrix) are rather small, ARSE is only slightly faster than RSE; however, as the dataset gets larger (e.g., \cref{fig:Star11_S00075_O50}), ARSE becomes significantly faster than RSE.

Regarding the accuracy of both methods, RSE and ARSE are virtually equal for the examples in \cref{fig:test_2d}, except in \cref{fig:Stairs_S0015_060,fig:Stairs4_S00075_O60}, where RSE performs slightly better.
We also observe in \cref{fig:test_2d,fig:test_2d_circles} that the proposed approach can correctly recover overlapping models. This is an intrinsic limitation of previous state-of-the-art competitors such as J-linkage~\cite{toldo08}, T-linkage~\cite{magri2014}, RPA~\cite{magri2015} and most multiple model estimation techniques, which are generally based on partitioning (clustering) the dataset.

\begin{table}
	\caption{Comparative performance of RSE and ARSE for 2D lines and 2D circles detection. The datasets (\cref{fig:test_2d,fig:test_2d_circles}) were created \cite{toldo08} by sampling points from different parametric models, adding a certain amount of noise to these points, and then further sampling outliers from a uniform distribution. Both methods, RSE and ARSE, perform well on these datasets. RSE results are slightly better, while ARSE is faster.}
	\label{tab:test_2d}
	
	\newcolumntype{Y}{S[table-format=4.0]}
	\newcolumntype{Z}{S[table-format=2.2]}
	\newcolumntype{W}{S[table-format=1.3]}
	\tabucolumn{Y}
	\tabucolumn{Z}
	\tabucolumn{W}
	
	\begin{subtable}{\textwidth}
		\caption{The original ground truth does not consider model intersections (overlaps) nor ``false'' outliers that are actually located very near some model. This issue artificially affects the methods' precision.}
		\label{tab:test_2d_given_gt}
		
		\centering
		\begin{small}
			\begin{tabu}{l *{2}{Y} *{2}{Z} *{6}{W}}
				\toprule
				& \multicolumn{2}{c}{Pref.~matrix} & \multicolumn{2}{c}{Time (\si{\second})} & \multicolumn{2}{c}{GNMI} & \multicolumn{2}{c}{Precision} & \multicolumn{2}{c}{Recall} \\
				\cmidrule(lr){2-3} \cmidrule(lr){4-5} \cmidrule(lr){6-7} \cmidrule(lr){8-9} \cmidrule(lr){10-11}
				& {m} & {n} & {RSE} & {ARSE} & {RSE} & {ARSE} & {RSE} & {ARSE} & {RSE} & {ARSE}  \\
				\midrule
				
				\cref{fig:Star5_S0_O5}         & 250  & 1277 & 3.26  & 1.97 & 0.770 & 0.774 & 0.851 & 0.854 & 1.000 & 1.000 \\
				\cref{fig:Star5_S0015_O0}      & 250  & 1004 & 3.34  & 2.17 & 0.728 & 0.722 & 0.837 & 0.832 & 0.984 & 0.984 \\
				\cref{fig:Star5_S0015_O50}     & 500  & 692  & 3.32  & 1.68 & 0.678 & 0.634 & 0.723 & 0.698 & 1.000 & 0.984 \\
				\cref{fig:Star5_S00075_O0}     & 250  & 1133 & 3.06  & 1.93 & 0.794 & 0.785 & 0.867 & 0.861 & 1.000 & 1.000 \\
				\cref{fig:Star5_S00075_O50}    & 500  & 687  & 3.70  & 1.48 & 0.683 & 0.686 & 0.729 & 0.732 & 1.000 & 1.000 \\
				\cref{fig:Star5_S00075_O75}    & 750  & 557  & 3.53  & 1.53 & 0.639 & 0.636 & 0.655 & 0.651 & 1.000 & 1.000 \\
				\cref{fig:Star11_S00075_O0}    & 550  & 1080 & 13.17 & 4.16 & 0.698 & 0.697 & 0.759 & 0.761 & 0.987 & 0.984 \\
				\cref{fig:Star11_S00075_O50}   & 1100 & 668  & 14.13 & 3.60 & 0.651 & 0.644 & 0.657 & 0.654 & 0.993 & 0.989 \\
				\cref{fig:Stairs_S0015_060}    & 500  & 917  & 6.53  & 2.05 & 0.639 & 0.576 & 0.699 & 0.654 & 0.990 & 0.970 \\
				\cref{fig:Stairs4_S00075_O60}  & 500  & 1153 & 7.72  & 2.04 & 0.619 & 0.604 & 0.660 & 0.644 & 1.000 & 1.000 \\
				\cref{fig:test_2d_circles} & 500  & 144  & 1.43  & 1.00 & 0.627 & 0.627 & 0.666 & 0.666 & 1.000 & 1.000 \\
				\midrule
				Mean                  &      &      & 5.74  & 2.15 & 0.684 & 0.671 & 0.737 & 0.728 & 0.996 & 0.992 \\
				STD                   &      &      & 4.27  & 0.93 & 0.059 & 0.068 & 0.081 & 0.086 & 0.006 & 0.010 \\
				Median                &      &      & 3.53  & 1.97 & 0.678 & 0.644 & 0.723 & 0.698 & 1.000 & 1.000	\\
				\bottomrule 
				
			\end{tabu}
		\end{small}
	\end{subtable}
	\par\bigskip
	
	\begin{subtable}{\textwidth}
		\caption{We reestimate the ground truth by accounting for intersections and ``false'' outliers. As we can see, the precision is now as high as the recall.}
		\label{tab:test_2d_restimated_gt}
		
		\centering
		\begin{small}
			\begin{tabu}{l *{2}{Y} *{6}{W}}
				\toprule
				& \multicolumn{2}{c}{Pref.~matrix} & \multicolumn{2}{c}{GNMI} & \multicolumn{2}{c}{Precision} & \multicolumn{2}{c}{Recall} \\
				\cmidrule(lr){2-3} \cmidrule(lr){4-5} \cmidrule(lr){6-7} \cmidrule(lr){8-9}
				& {m} & {n} & {RSE} & {ARSE} & {RSE} & {ARSE} & {RSE} & {ARSE}  \\
				\midrule
				
				\cref{fig:Star5_S0_O5}         & 250  & 1277 & 0.984 & 0.976 & 0.997 & 0.997 & 0.997 & 0.993 \\
				\cref{fig:Star5_S0015_O0}      & 250  & 1004 & 0.933 & 0.919 & 0.981 & 0.975 & 0.987 & 0.987 \\
				\cref{fig:Star5_S0015_O50}     & 500  & 692  & 0.968 & 0.852 & 0.986 & 0.939 & 0.997 & 0.968 \\
				\cref{fig:Star5_S00075_O0}     & 250  & 1133 & 0.985 & 0.928 & 0.993 & 0.976 & 1.000 & 0.990 \\
				\cref{fig:Star5_S00075_O50}    & 500  & 687  & 0.962 & 0.945 & 0.986 & 0.982 & 0.995 & 0.986 \\
				\cref{fig:Star5_S00075_O75}    & 750  & 557  & 0.961 & 0.905 & 0.990 & 0.968 & 0.990 & 0.972 \\
				\cref{fig:Star11_S00075_O0}    & 550  & 1080 & 0.974 & 0.956 & 0.992 & 0.99  & 0.994 & 0.986 \\
				\cref{fig:Star11_S00075_O50}   & 1100 & 668  & 0.950 & 0.945 & 0.980 & 0.977 & 0.989 & 0.988 \\
				\cref{fig:Stairs_S0015_060}    & 500  & 917  & 0.819 & 0.633 & 0.943 & 0.838 & 0.946 & 0.880 \\
				\cref{fig:Stairs4_S00075_O60}  & 500  & 1153 & 0.839 & 0.705 & 0.926 & 0.862 & 0.955 & 0.913 \\
				\cref{fig:test_2d_circles} & 500  & 144  & 0.936 & 0.936 & 0.979 & 0.979 & 0.988 & 0.988 \\
				\midrule
				Mean                  &                   &      & 0.937 & 0.882 & 0.978 & 0.953 & 0.985 & 0.968 \\
				STD                   &                   &      & 0.056 & 0.111 & 0.022 & 0.053 & 0.018 & 0.037 \\
				Median                &                   &      & 0.961 & 0.928 & 0.986 & 0.976 & 0.990 & 0.986 \\
				\bottomrule
			\end{tabu}
		\end{small}		
	\end{subtable}
	
\end{table}

\paragraph{Vanishing points.}
Under the assumption of perspective projection (e.g., with a pin-hole camera), sets of parallel lines in 3D space are projected into a 2D image as a set of concurrent lines. The point in the image plane at which these lines meet is called a \textit{vanishing point}. Vanishing points provide important information to make inferences about the 3D structures of a scene from its 2D (projected) image. Since these projections are non-invertible, concurrence in the image plane does not necessarily imply parallelism in 3D. This said, the counterexamples for this implication are rare in real images, and the problem of finding parallel lines in 3D is reduced in practice to finding vanishing points in the image plane~\cite[e.g.,][]{almansa03vp,Andalo2012,denis08,lezama14,tardif09}. We are here interested in finding vanishing points when no a priori information is known about the image or the camera with which it was taken.

The dataset elements in this case are line segments extracted with a suitable algorithm (e.g.,~\cite{grompone10}).
As the distance between a vanishing point and a line segment, we consider the angular difference between the line segment and the line joining the vanishing point and the segment's midpoint.

To evaluate the proposed framework for the task of detecting vanishing points, we use the York Urban database~\cite{denis08}, which comes with ground truth assignments of image segments to three orthogonal directions in 3D.
Of course, since the dataset images represent urban scenes, performance can be readily boosted by adding problem-specific constraints to the proposed algorithm as an extra postprocessing step; common simplifications include knowledge about the camera calibration and the Manhattan world assumption~\cite[e.g.,][]{lezama14}.

\begin{figure}[t]
	\centering
	\begin{small}
	\begin{tabu} to \textwidth {@{\hspace{0cm}} *{3}{X[c,m] @{\hspace{.15cm}}} X[c,m] @{\hspace{0cm}} }
		\fbox{\includegraphics[width=.99\linewidth]{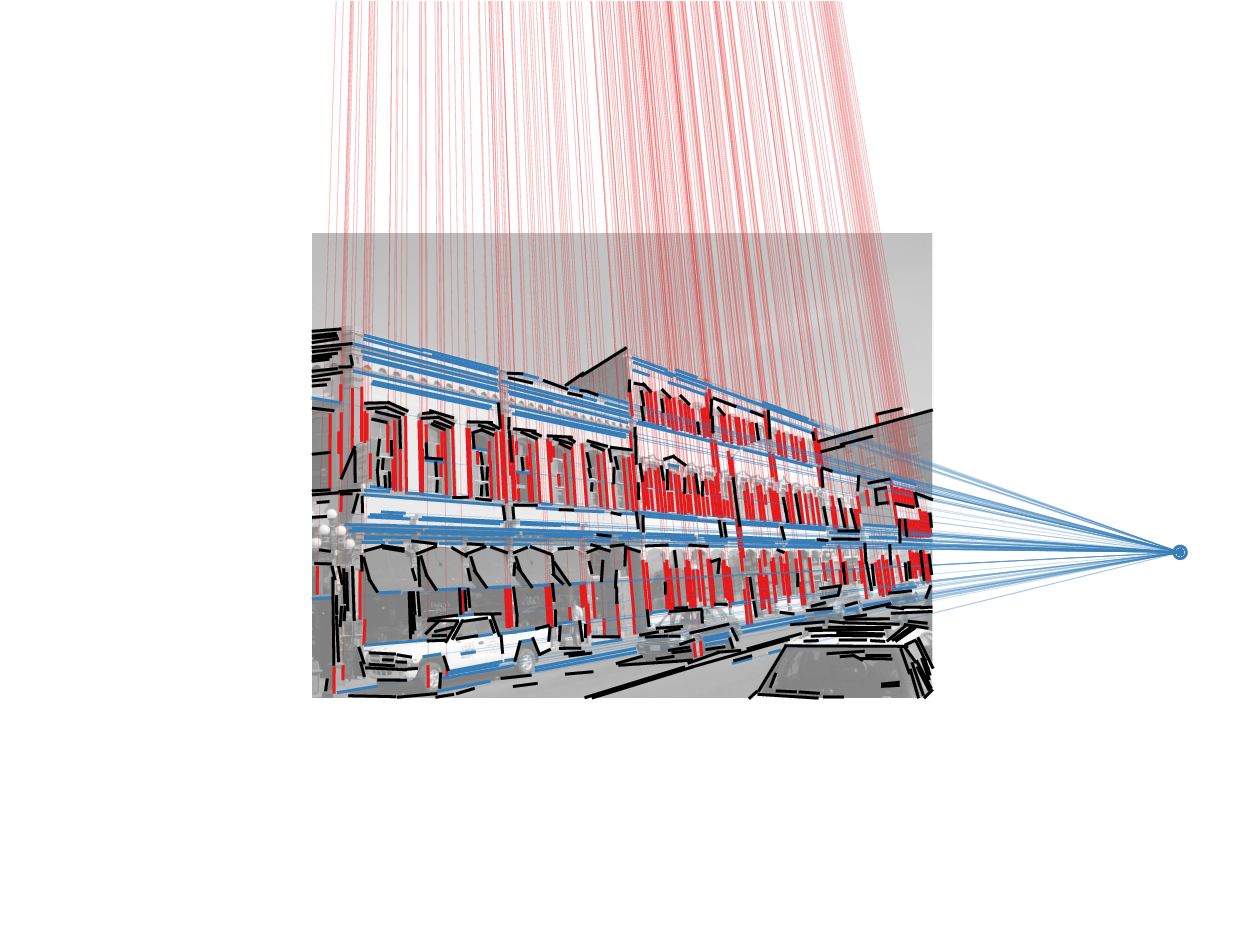}} &
		\fbox{\includegraphics[width=.99\linewidth]{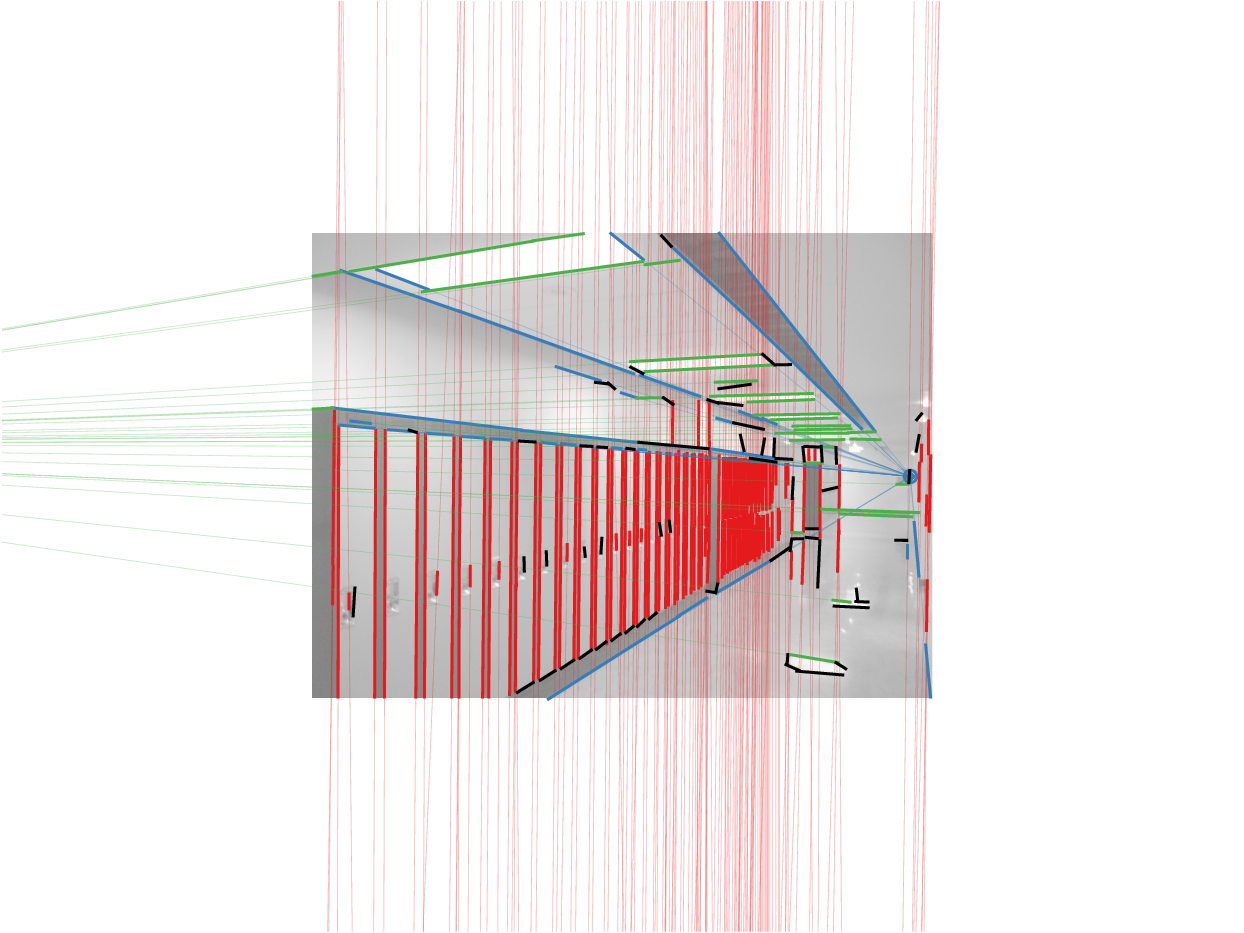}} &
		\fbox{\includegraphics[width=.99\linewidth]{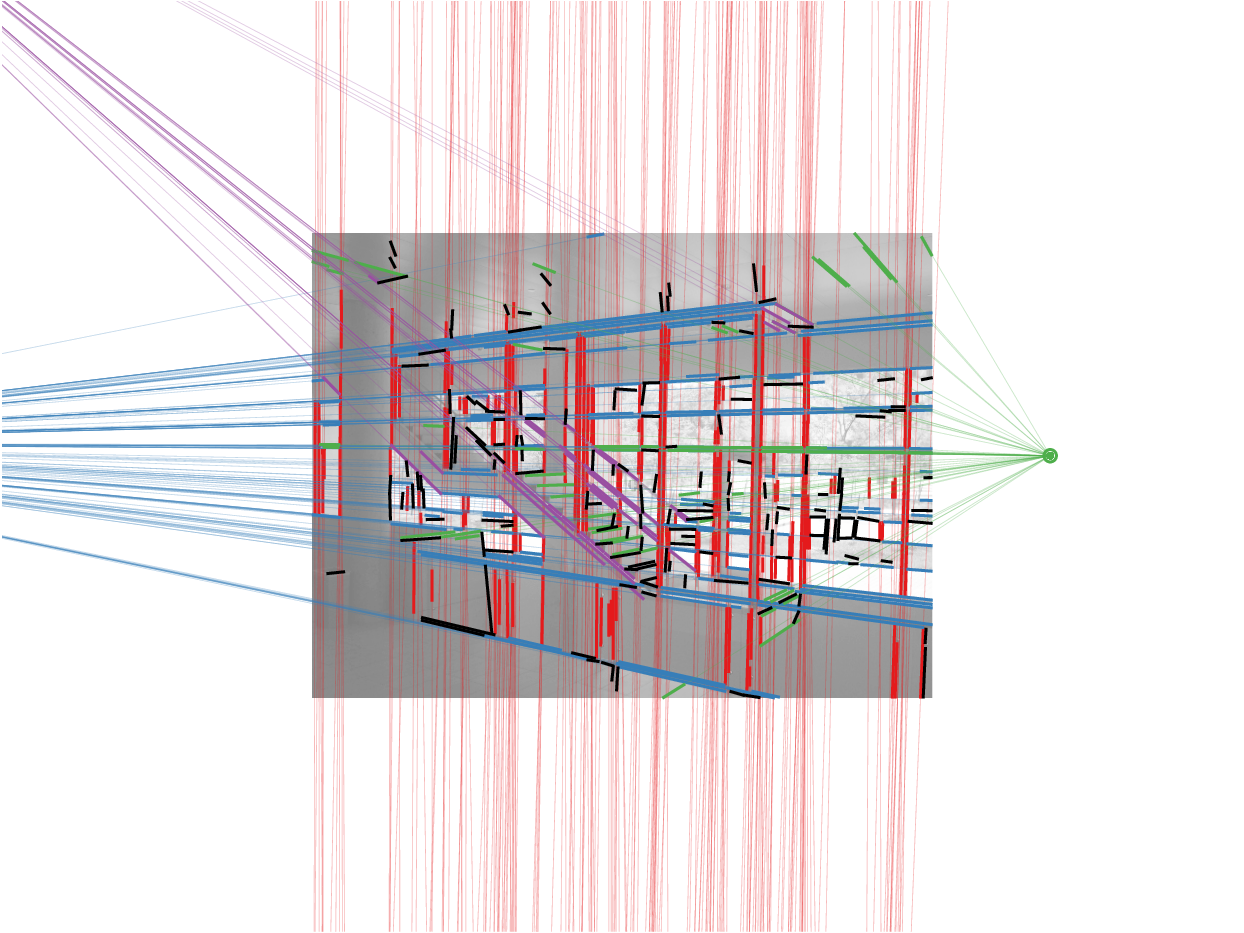}} &		
		\fbox{\includegraphics[width=.99\linewidth]{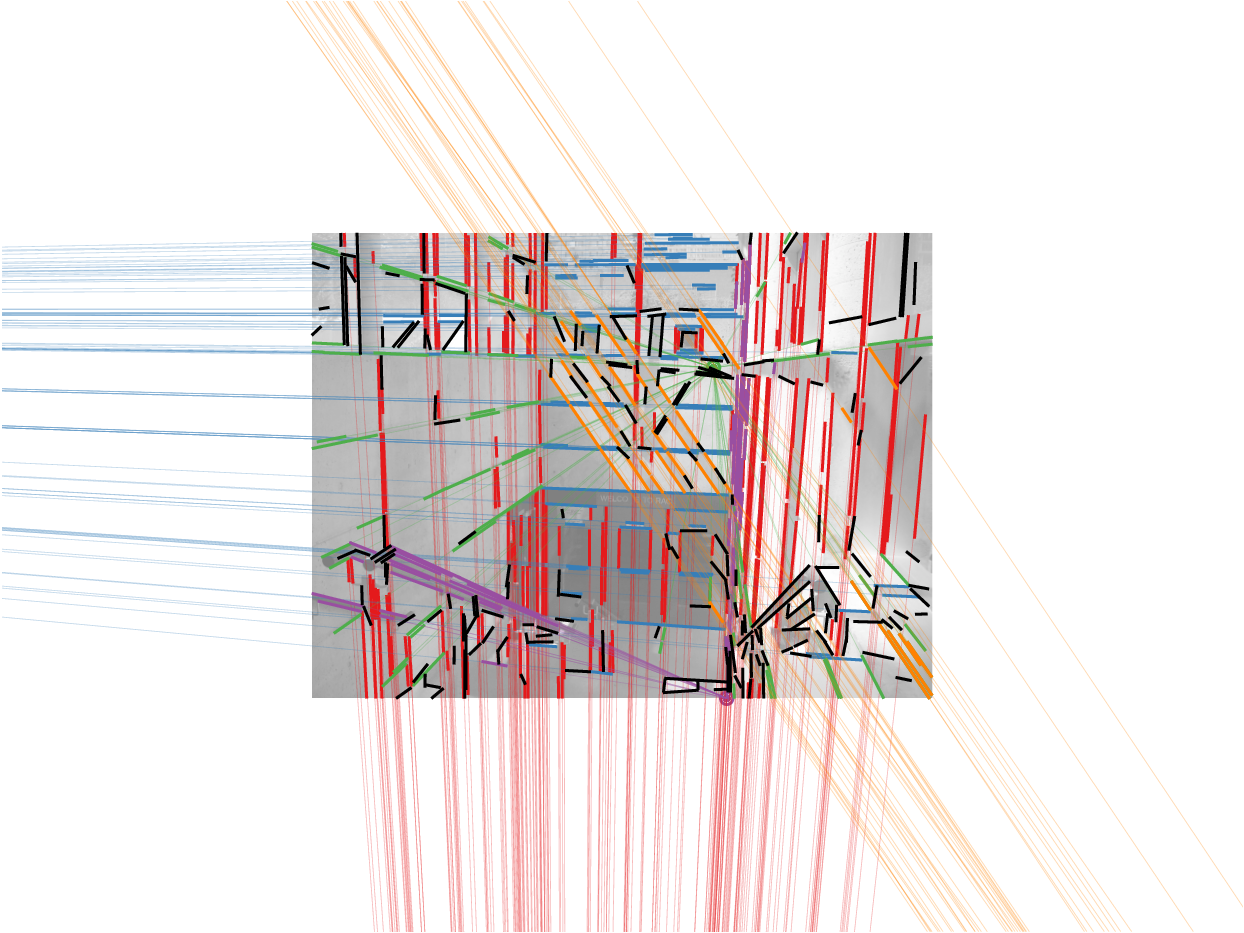}} \\
	\end{tabu}
	\end{small}	
	
	\caption{Examples of vanishing points detection on the York Urban database~\cite{denis08}. ARSE can detect any number of vanishing points (from left to right, 2, 3, 4, and 5 detections, respectively) , even when their directions are not orthogonal in 3D.  For this experiment, we set $\delta = 10^{-2} \pi$.}
	\label{fig:test_vp}
\end{figure}

The results of the proposed method are presented in \cref{tab:test_vp}. There we can observe that RSE and ARSE perform well in terms of precision and recall. ARSE is in average twice as fast as RSE. In \cref{fig:test_vp} we can observe a few characteristic examples of ARSE's results. Notice that in the two rightmost examples, the method correctly finds more than three vanishing points; these additional points are deemed incorrect in the results of \cref{tab:test_vp}, as the ground truth only contemplates three points.

\begin{table}[t]
	\caption{Comparative performance of RSE and ARSE to detect vanishing points on the York Urban database~\cite{denis08}. The ground truth assumes a Manhattan world; hence, it only contains 2 or 3 vanishing points per image. This artificially affects the performance of the proposed method, since some images actually contain more than 3 vanishing points (see \cref{fig:test_vp}).}
	\label{tab:test_vp}
	
	\newcolumntype{Z}{S[table-format=1.2]}
	\newcolumntype{W}{S[table-format=1.3]}
	\tabucolumn{Z}
	\tabucolumn{W}
	
	\centering
	\begin{small}
		\begin{tabu}{l *{2}{Z} *{6}{W}}
			\toprule
			& \multicolumn{2}{c}{Time (\si{\second})} & \multicolumn{2}{c}{GNMI} & \multicolumn{2}{c}{Precision} & \multicolumn{2}{c}{Recall} \\
			\cmidrule(lr){2-3} \cmidrule(lr){4-5} \cmidrule(lr){6-7} \cmidrule(lr){8-9}
			& {RSE} & {ARSE} & {RSE} & {ARSE} & {RSE} & {ARSE} & {RSE} & {ARSE}  \\
			\midrule
			Mean	& 3.66	& 1.52		& 0.735	& 0.695	& 0.853	& 0.833	& 0.868	& 0.852 \\
			STD		& 6.02   	& 0.69      & 0.155	& 0.151	& 0.161	& 0.155	& 0.149	& 0.139 \\
			Median	& 2.09    & 1.42		& 0.764		& 0.706	& 0.939	& 0.916	& 0.938	& 0.915 \\
			\bottomrule
		\end{tabu}
	\end{small}
	
\end{table}

\paragraph{Fundamental matrices and homographies.}

\begin{figure}[t]
	
	\centering
	\begin{tabu} to .85\textwidth {*{2}{X[c,m]}}
		\includegraphics[width=\linewidth]{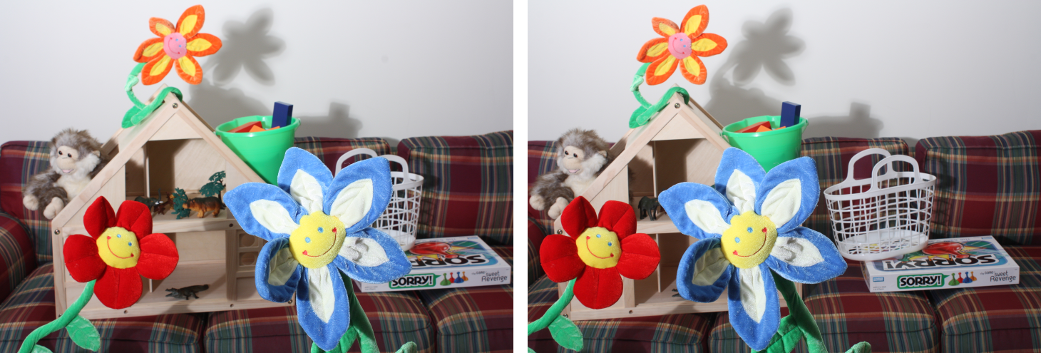} &
		\includegraphics[width=\linewidth]{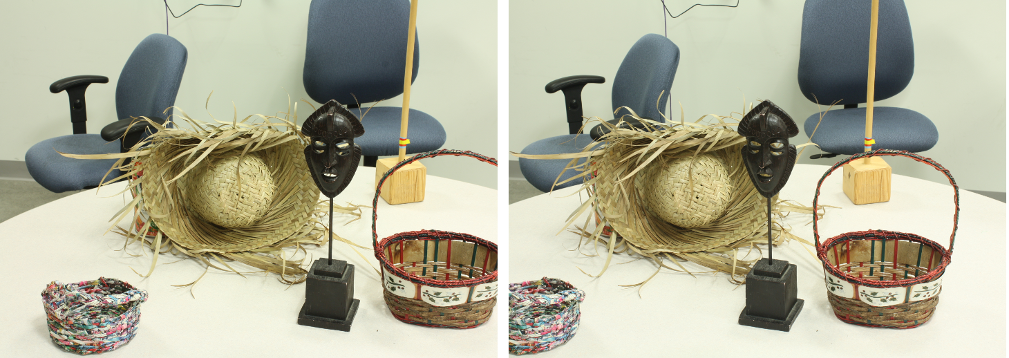} \\
		\\[-1.5ex]
		\includegraphics[width=\linewidth]{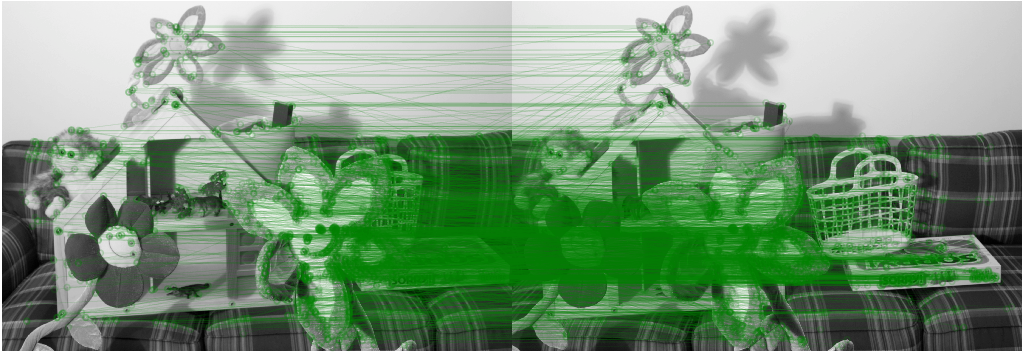} &
		\includegraphics[width=\linewidth]{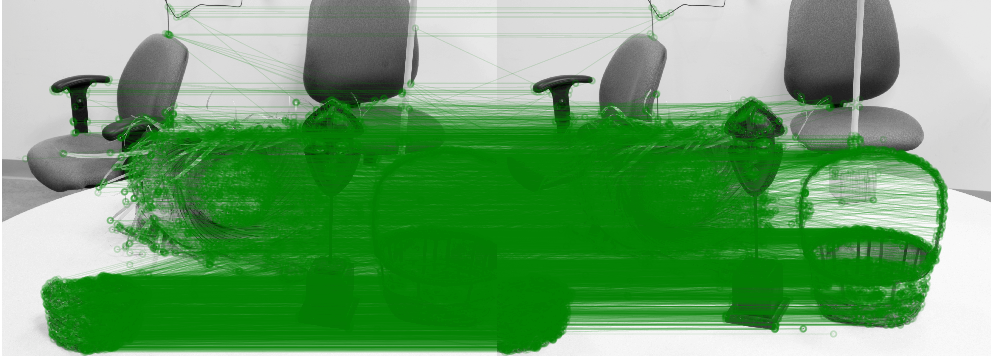} \\
		\\[-1.5ex]
		\includegraphics[width=\linewidth]{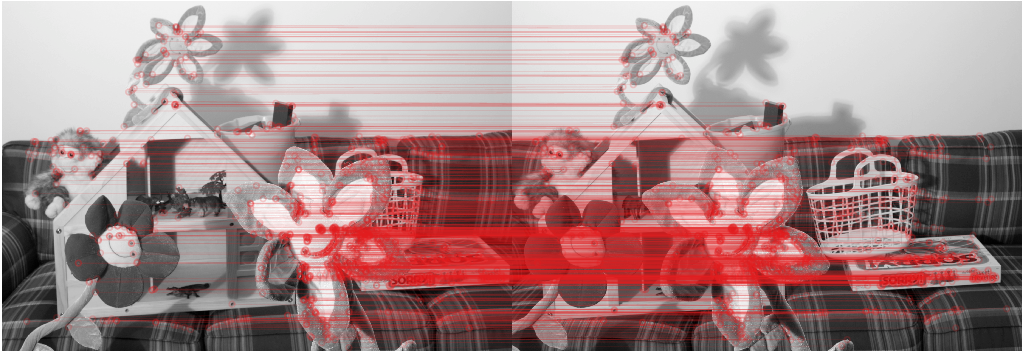} &
		\includegraphics[width=\linewidth]{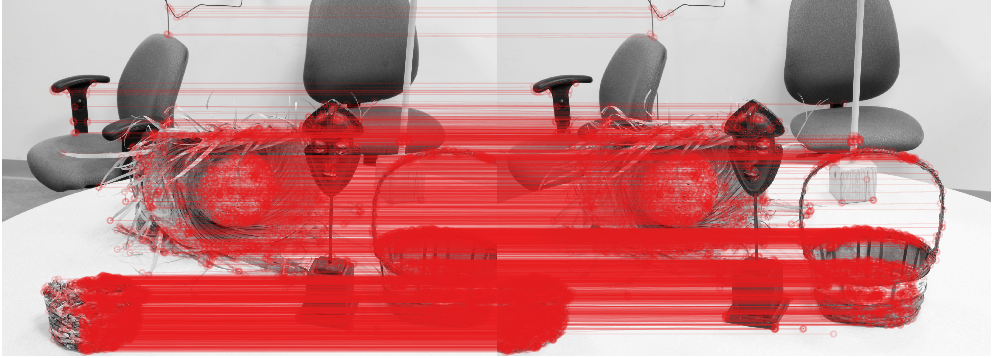} \\
		\\[-1.5ex]
		\includegraphics[width=\linewidth]{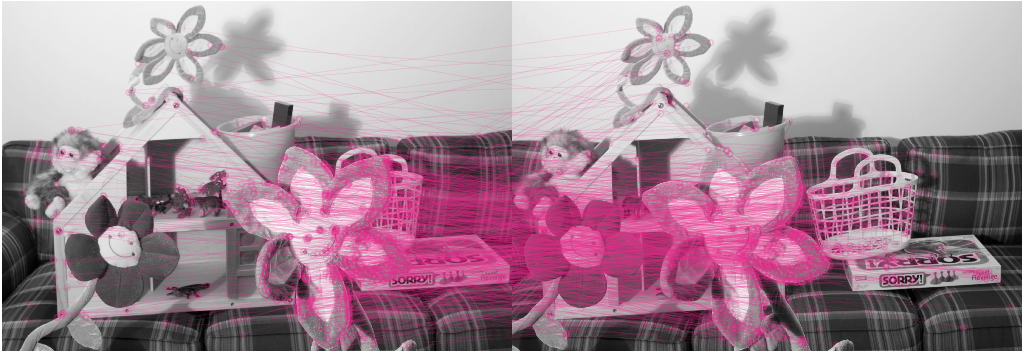} &
		\includegraphics[width=\linewidth]{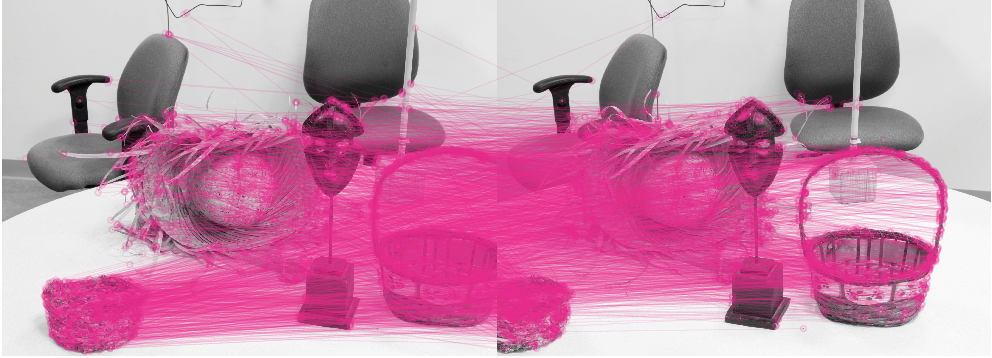} \\
	\end{tabu}
				
	\caption{Fundamental matrix estimation on two stereo pairs of the Middlebury Stereo dataset~\cite{Scharstein2014}.
		The stereo pairs are calibrated such that all epipolar lines are horizontal.
		Since the scene is rigid, a single fundamental matrix is enough to describe the stereo geometry. In both cases, a single model (fundamental matrix) was found by both ARSE and RSE.
		From top to bottom: original pair, matches obtained from BRISK keypoints  and descriptors \cite{Leutenegger2011} ($2518$ on the left, $8999$ on the right), ARSE inliers ($1363/2518 \approx 54\%$ on the left, $5537/8999 \approx 61\%$ on the right), and ARSE outliers (the RSE inlier sets are very similar). As expected, the lines connecting inlier matches are horizontal.}
	\label{fig:middlebury}	
\end{figure}

For these examples, where we are trying to estimate geometric transforms between two images, the base elements are keypoint matches. Any modern method to find those matches would work relatively well for our purposes; we use BRISK \cite{Leutenegger2011}.
We show in \cref{fig:middlebury} two examples of fundamental matrix estimation on a stereo pair of images. The only motion in the scene corresponds to a change in camera perspective, and it can be described with a single fundamental matrix.
Both methods achieve extremely similar results and in both cases (left and right pairs) they find a single model.
The example on the left contains 2518 matches. ARSE detects 1377 matches in $7.99$ seconds, while RSE detects 1363 matches in $98.32$ seconds.
ARSE's precision and recall, considering the result of RSE as ground truth, are $0.959$ and $0.968$, respectively.
The example on the right contains 8999 matches. ARSE detects 5533 matches in $30.40$ seconds, whereas RSE detects 5537 matches in $1033.45$ seconds.
In this case, the precision is $0.959$ and recall is $0.968$. However, there is a huge difference in speed between both methods, which is accentuated with the size of the dataset. On the left example, ARSE is about 12 times faster, whereas on the right example it is about 34 times faster.

It is important to emphasize that other random sampling techniques can be used beyond the baseline in \cref{algo:random_sample}. To show this, we have included experiments using the state-of-the-art MultiGS algorithm~\cite{chin2012}. It is worth pointing out that a recent technique~\cite{BenArtzi2016} has been able to reduce the size of the minimal sample set for fundamental matrices to $b=2$; its use would significantly reduce the computational complexity of the sampling step, imposing an upper bound of $O(m^2)$ to the total number of possible combinations.

\begin{figure}[t]

	\begin{subfigure}{\textwidth}
		\centering
		\includegraphics[height=.17\linewidth]{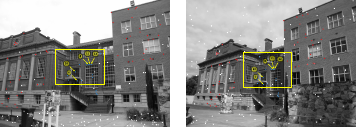}%
		\hfill
		\includegraphics[height=.17\linewidth]{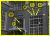}
		\includegraphics[height=.17\linewidth]{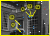}

		\caption{Homography: the ``barrsmith'' pair contains 5 errors out of 235 matches = 2.35\%}
	\end{subfigure}

	\begin{subfigure}{\textwidth}
		\centering
		\includegraphics[height=.17\linewidth]{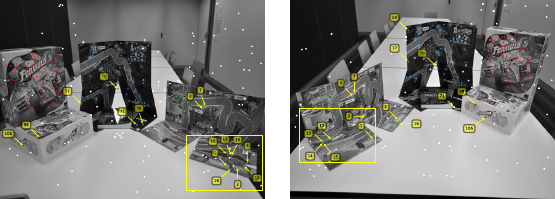}%
		\hfill
		\includegraphics[height=.17\linewidth]{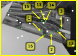}		
		\includegraphics[height=.17\linewidth]{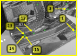}
		
		\caption{Fundamental matrix: the ``boardgame'' pair contains 16 errors out of 266 matches = 6.02\%}
	\end{subfigure}
	
	\caption{The ground truth of the standard AdelaideRMF dataset~\cite{wong11} has non-negligible errors (detectable by simple ocular inspection). We provide two examples of such errors, one for each type of geometric model. On each example, we show the original pair on the left, and provide a zoom-in to a specific region on the right. Inliers are indicated in color and outliers in white; the wrong matches are identified with matching numbers.}
	\label{fig:adelaide_errors}	
\end{figure}

We estimate multiple fundamental matrices (moving camera and moving objects) and multiple homographies (moving planar objects) on the images in the AdelaideRMF dataset \cite{wong11}. Although we use this dataset because it is standard in the literature \cite[e.g.,][]{magri2014,magri2015,Pham2014,wong11}, we would like to point out that the ground truth of this dataset contains a non-negligible quantity of severe errors. In \cref{fig:adelaide_errors} we show two examples where these errors are easily identified upon visual inspection. The dataset's ground truth would need a thorough revision to be truly useful as a scientific benchmark. We include these results nonetheless for the sake of completeness.

\begin{figure}[t]
	\begin{subfigure}{\textwidth}
		\includegraphics[width=\linewidth]{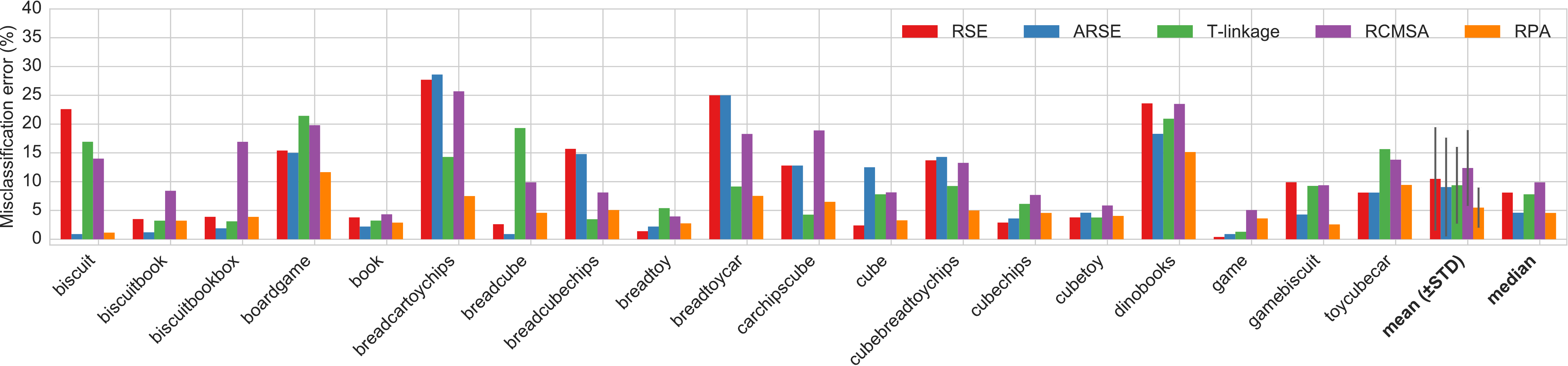}
		\caption{Fundamental matrices: $\delta = 3$}
		\label{fig:fundamental}
	\end{subfigure}
	
	\begin{subfigure}{\textwidth}
		\includegraphics[width=\linewidth]{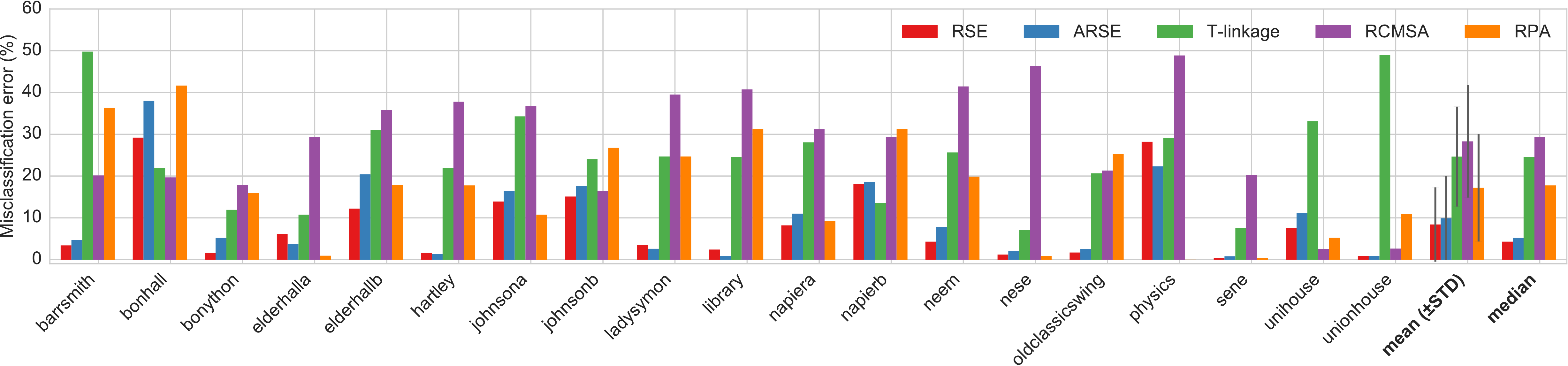}
		\caption{Homographies: $\delta = 14.5$}
		\label{fig:homography}
	\end{subfigure}
	
	\caption{Results on the AdelaideRMF dataset~\cite{wong11} for the estimation of fundamental matrices and homographies. Note that RPA does not automatically estimate the number of models (the ground truth number of models was used in all examples). In each subfigure, we indicate the value chosen for $\delta$ (\cref{eq:consensusSet}).}
	\label{fig:transforms}
\end{figure}

The results on the full dataset are presented in \cref{fig:transforms}. We include comparisons with T-linkage~\cite{magri2014}, RCMSA~\cite{Pham2014}, and RPA~\cite{magri2015}. Among these three methods, RPA is the one that achieves the best performances; however, to yield these performances RPA uses as an input the \emph{ground truth} number of models to estimate. The other methods, as the proposed method, automatically estimate this number.

In \cref{fig:fundamental}, RSE and ARSE perform on average similarly as T-linkage. However, notice that the median performance of ARSE is similar to the one of RPA, with the added benefit of not having to tune a critical and fundamental parameter. In \cref{fig:homography}, both RSE and ARSE clearly outperform all other methods. The complete numerical results are included in \cref{tab:fundamental,tab:homography}.
	
\begin{table}[t]
	\centering
	\caption{Results on the AdelaideRMF dataset~\cite{wong11} for the estimation of multiple fundamental matrices, see \cref{fig:fundamental}.}
	\label{tab:fundamental}
	
	\begin{footnotesize}
		\begin{tabu}{l *{7}{S[table-format=2.2,round-integer-to-decimal=true,round-mode=places]}}
			\toprule
			& \multicolumn{2}{c}{Time (\si{\second})} & \multicolumn{5}{c}{Misclassification error} \\
			\cmidrule(lr){2-3} \cmidrule(lr){4-8}
			& {RSE} & {ARSE} & {RSE} & {ARSE} & {T-linkage~\cite{magri2014}} & {RCMSA~\cite{Pham2014}} & {RPA~\cite{magri2015}}  \\
			\midrule
			
			biscuit           & 2.18    & 1.23       & 22.60                   & 0.90       & 16.93     & 14    & 1.15  \\
			biscuitbook       & 3.09    & 1.31       & 3.50                    & 1.20       & 3.23      & 8.41  & 3.23  \\
			biscuitbookbox    & 2.34    & 0.99       & 3.90                    & 1.90       & 3.1       & 16.92 & 3.88  \\
			boardgame         & 1.76    & 0.56       & 15.40                   & 15.00      & 21.43     & 19.8  & 11.65 \\
			book              & 4.09    & 1.09       & 3.80                    & 2.20       & 3.24      & 4.32  & 2.88  \\
			breadcartoychips  & 0.39    & 0.29       & 27.70                   & 28.60      & 14.29     & 25.69 & 7.5   \\
			breadcube         & 4.13    & 1.42       & 2.60                    & 0.90       & 19.31     & 9.87  & 4.58  \\
			breadcubechips    & 1.2     & 0.89       & 15.70                   & 14.80      & 3.48      & 8.12  & 5.07  \\
			breadtoy          & 5.75    & 1.55       & 1.40                    & 2.20       & 5.4       & 3.96  & 2.76  \\
			breadtoycar       & 0.16    & 0.17       & 25.00                   & 25.00      & 9.15      & 18.29 & 7.52  \\
			carchipscube      & 1.1     & 0.46       & 12.80                   & 12.80      & 4.27      & 18.9  & 6.5   \\
			cube              & 1.12    & 0.63       & 2.40                    & 12.50      & 7.8       & 8.14  & 3.28  \\
			cubebreadtoychips & 1.56    & 1.02       & 13.70                   & 14.30      & 9.24      & 13.27 & 4.99  \\
			cubechips         & 1.11    & 0.57       & 2.90                    & 3.60       & 6.14      & 7.7   & 4.57  \\
			cubetoy           & 2.11    & 0.87       & 3.80                    & 4.60       & 3.77      & 5.86  & 4.04  \\
			dinobooks         & 2.75    & 0.96       & 23.60                   & 18.30      & 20.94     & 23.5  & 15.14 \\
			game              & 0.32    & 0.26       & 0.40                    & 0.90       & 1.3       & 5.07  & 3.62  \\
			gamebiscuit       & 2.09    & 0.87       & 9.90                    & 4.30       & 9.26      & 9.37  & 2.57  \\
			toycubecar        & 1.03    & 0.67       & 8.10                    & 8.10       & 15.66     & 13.81 & 9.43  \\
			
			\midrule
			
			Mean              & 2.01    & 0.83       & 10.48                   & 9.06       & 9.37      & 12.37 & 5.49  \\
			STD               & 1.45    & 0.39       & 8.97                    & 8.57       & 6.66      & 6.58  & 3.48  \\
			Median            & 1.76    & 0.87       & 8.10                    & 4.60       & 7.80      & 9.87  & 4.57 \\
			
			\bottomrule
		\end{tabu}
	\end{footnotesize}
\end{table}

\begin{table}[t]
	\centering
	\caption{Results on the AdelaideRMF dataset~\cite{wong11} for the estimation of multiple homographies, see \cref{fig:homography}.}
	\label{tab:homography}
	
	\begin{footnotesize}
		\begin{tabu}{l S[table-format=3.2,round-integer-to-decimal=true,round-mode=places] *{6}{S[table-format=2.2,round-integer-to-decimal=true,round-mode=places]}}
			\toprule
			& \multicolumn{2}{c}{Time (\si{\second})} & \multicolumn{5}{c}{Misclassification error} \\
			\cmidrule(lr){2-3} \cmidrule(lr){4-8}
			& {RSE} & {ARSE} & {RSE} & {ARSE} & {T-linkage~\cite{magri2014}} & {RCMSA~\cite{Pham2014}} & {RPA~\cite{magri2015}}  \\
			
			\midrule
			
			barrsmith       & 1.65    & 2.18       & 3.40                    & 4.70       & 49.79     & 20.14 & 36.31 \\
			bonhall         & 90.37   & 13.50      & 29.20                   & 38.00      & 21.84     & 19.69 & 41.67 \\
			bonython        & 1.93    & 1.60       & 1.60                    & 5.20       & 11.92     & 17.79 & 15.89 \\
			elderhalla      & 3.37    & 2.37       & 6.10                    & 3.70       & 10.75     & 29.28 & 0.93  \\
			elderhallb      & 6.08    & 4.86       & 12.20                   & 20.40      & 31.02     & 35.78 & 17.82 \\
			hartley         & 6.14    & 3.53       & 1.60                    & 1.30       & 21.90     & 37.78 & 17.78 \\
			johnsona        & 22.55   & 7.86       & 13.90                   & 16.40      & 34.28     & 36.73 & 10.76 \\
			johnsonb        & 81.41   & 0.93       & 15.10                   & 17.60      & 24.04     & 16.46 & 26.76 \\
			ladysymon       & 11.30   & 5.70       & 3.50                    & 2.60       & 24.67     & 39.50 & 24.67 \\
			library         & 4.03    & 2.66       & 2.40                    & 0.90       & 24.53     & 40.72 & 31.29 \\
			napiera         & 4.83    & 2.98       & 8.20                    & 11.00      & 28.08     & 31.16 & 9.25  \\
			napierb         & 11.34   & 4.88       & 18.10                   & 18.60      & 13.50     & 29.40 & 31.22 \\
			neem            & 9.51    & 2.73       & 4.30                    & 7.80       & 25.65     & 41.45 & 19.86 \\
			nese            & 13.05   & 6.21       & 1.20                    & 2.10       & 7.05      & 46.34 & 0.83  \\
			oldclassicswing & 15.60   & 6.07       & 1.70                    & 2.50       & 20.66     & 21.30 & 25.25 \\
			physics         & 2.93    & 3.38       & 28.20                   & 22.30      & 29.13     & 48.87 & 0.00  \\
			sene            & 8.22    & 4.03       & 0.40                    & 0.80       & 7.63      & 20.20 & 0.42  \\
			unihouse        & 229.62  & 17.10      & 7.60                    & 11.20      & 33.13     & 2.56  & 5.21  \\
			unionhouse      & 1.03    & 1.03       & 0.90                    & 0.90       & 48.99     & 2.64  & 10.87 \\
			
			\midrule
			
			Mean            & 27.63   & 4.93       & 8.40                    & 9.89       & 24.66     & 28.30 & 17.20 \\
			STD             & 55.02   & 4.14       & 8.90                    & 10.04      & 11.96     & 13.45 & 12.87 \\
			Median          & 8.22    & 3.53       & 4.30                    & 5.20       & 24.53     & 29.40 & 17.78 \\
			
			\bottomrule
		\end{tabu}
	\end{footnotesize}
\end{table}

\paragraph{3D planes estimation.}
In recent years, people have started ``scanning'' objects with range imaging hardware (e.g., Kinect or RealSense cameras) to obtain 3D point clouds. The geometric nature of man-made objects (composed of planes, cylinders, spheres, etc.) is very well adapted to the  analysis of their corresponding point clouds with MPME.

\begin{figure}[t]
	\begin{subfigure}{\textwidth}	
		\centering
		\begin{tabu}{@{\hspace{0cm}} *{5}{X[c,m] @{\hspace{0.1cm}}}}
			\includegraphics[width=\linewidth]{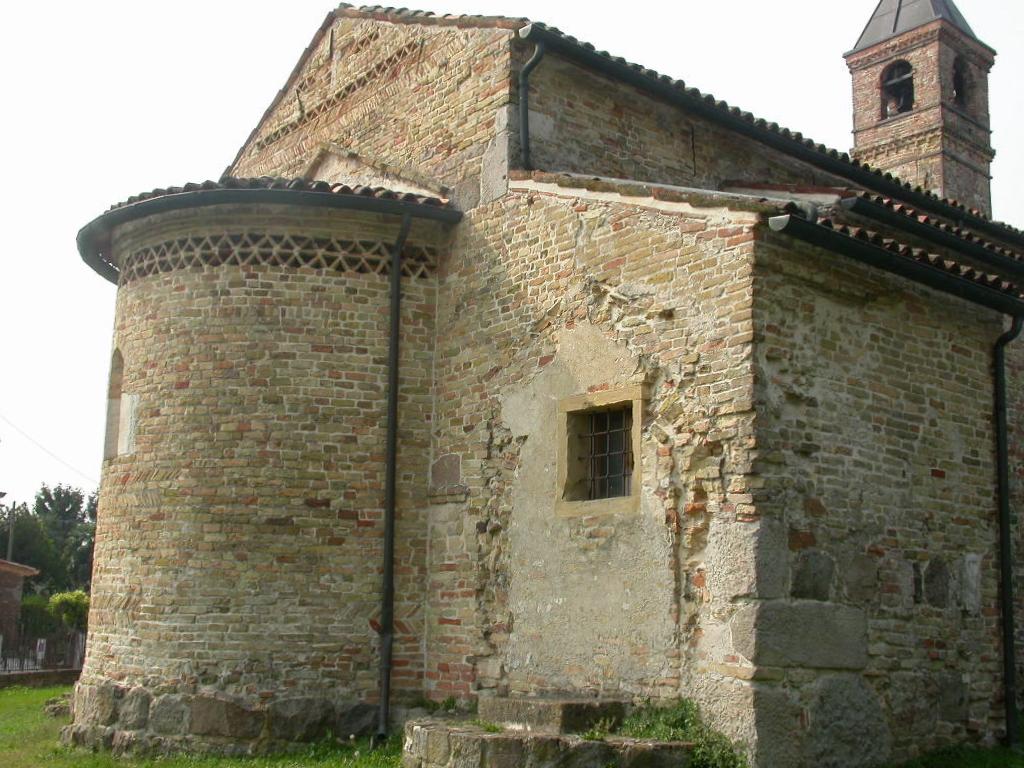} &
			\includegraphics[width=\linewidth]{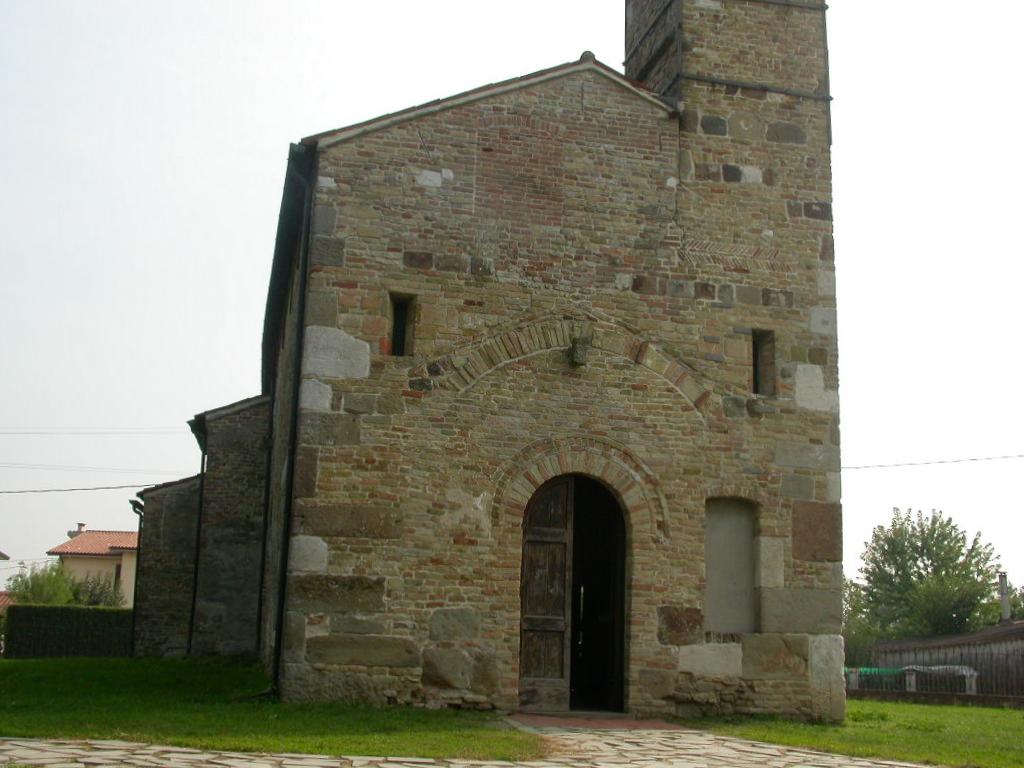} &
			\includegraphics[width=\linewidth]{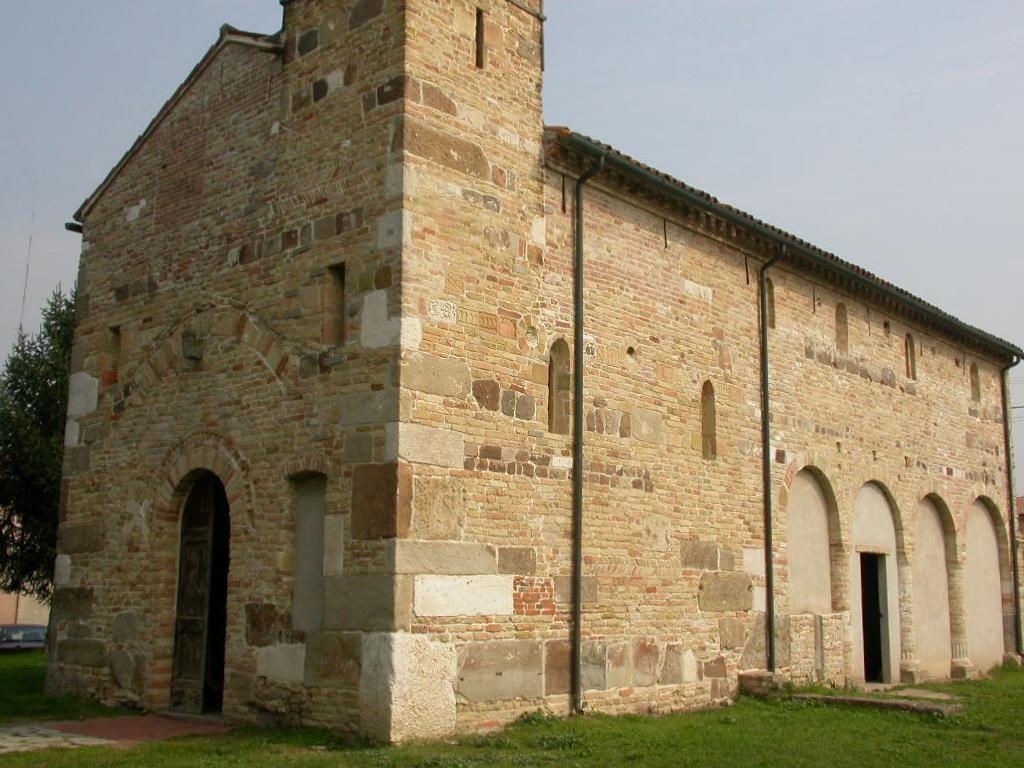} &
			\includegraphics[width=\linewidth]{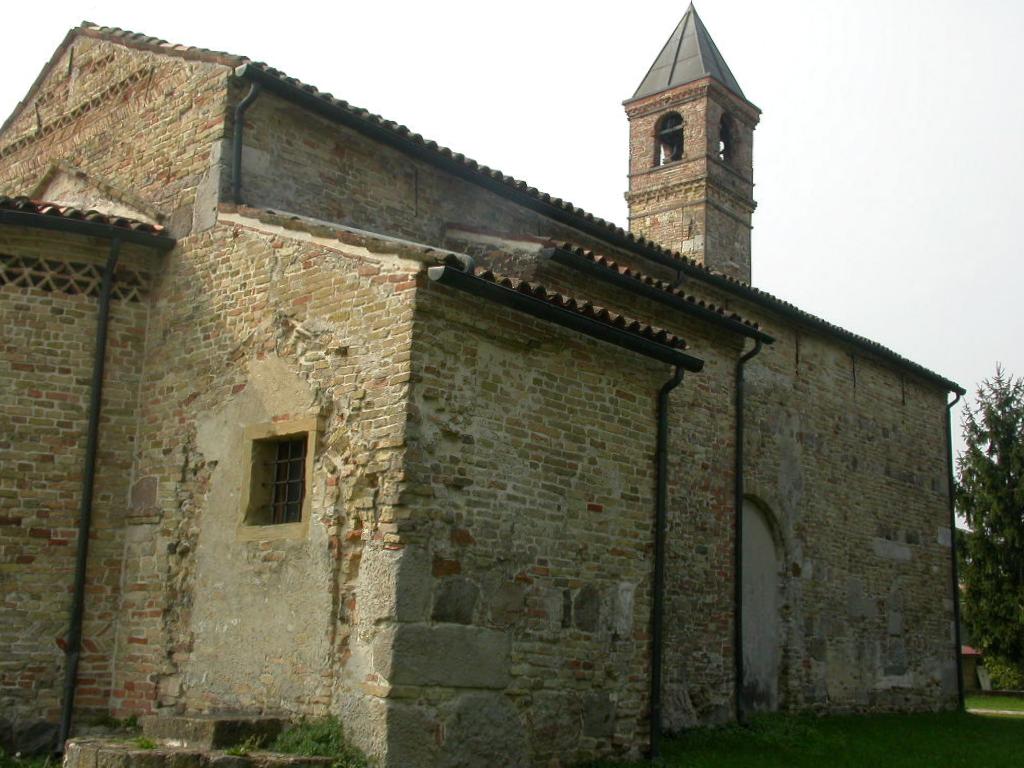} &
			\includegraphics[width=\linewidth]{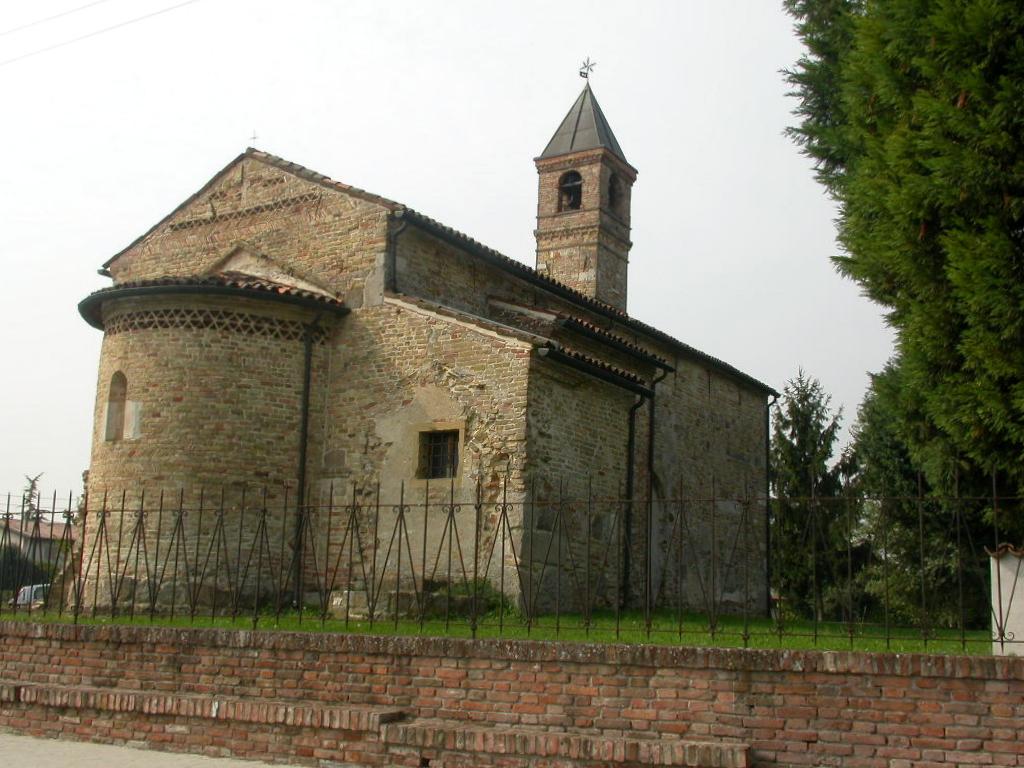} \\
		\end{tabu}
		\caption{A few samples of the original images used to build the 3D point cloud.}
		\label{fig:pozzoveggiani_images}
	\end{subfigure}
	
	\begin{subfigure}{\textwidth}
		\centering
		\begin{tabu}{@{\hspace{0cm}} *{4}{X[c,m] @{\hspace{0cm}}}}
			\includegraphics[width=\linewidth,trim={250px 180px 250px 100px},clip]{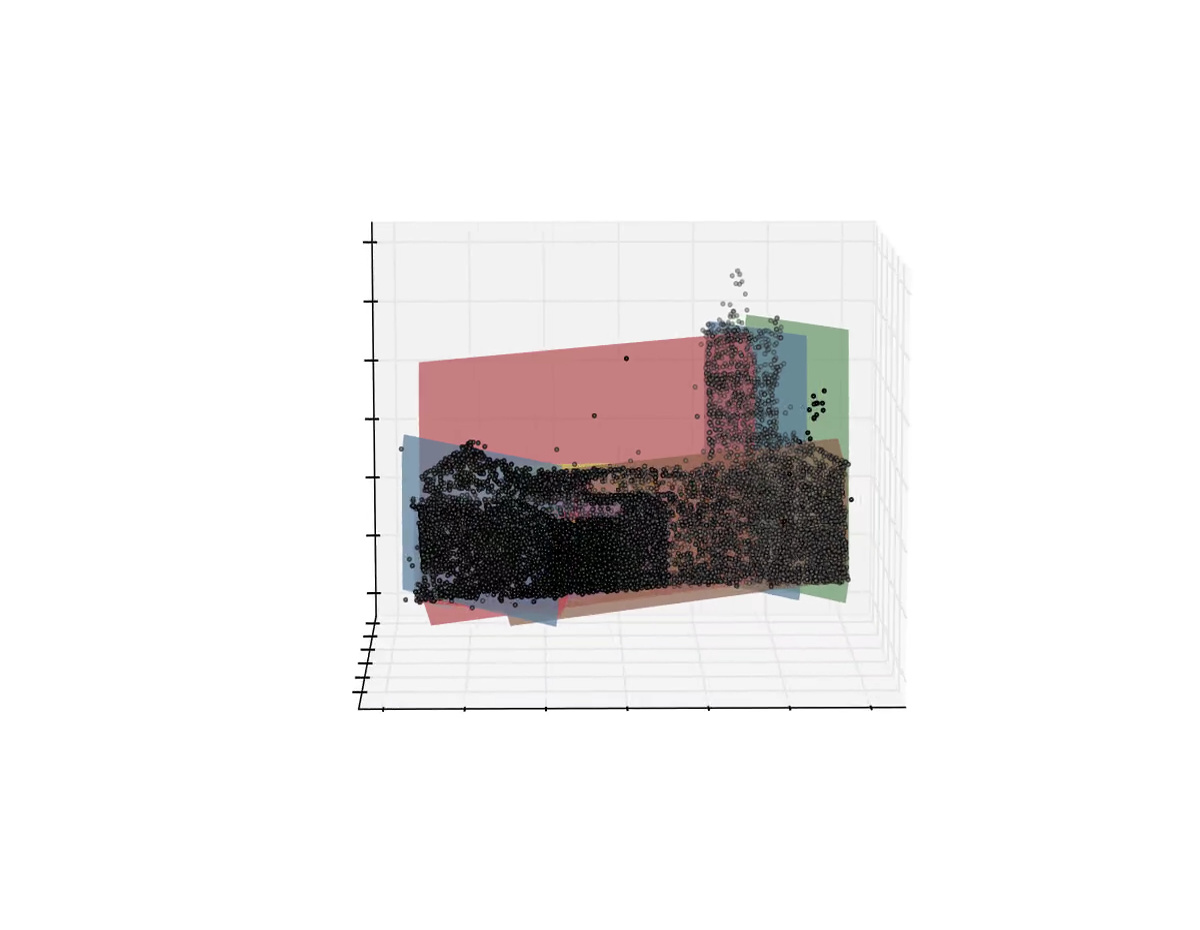} &
			\includegraphics[width=\linewidth,trim={250px 180px 250px 100px},clip]{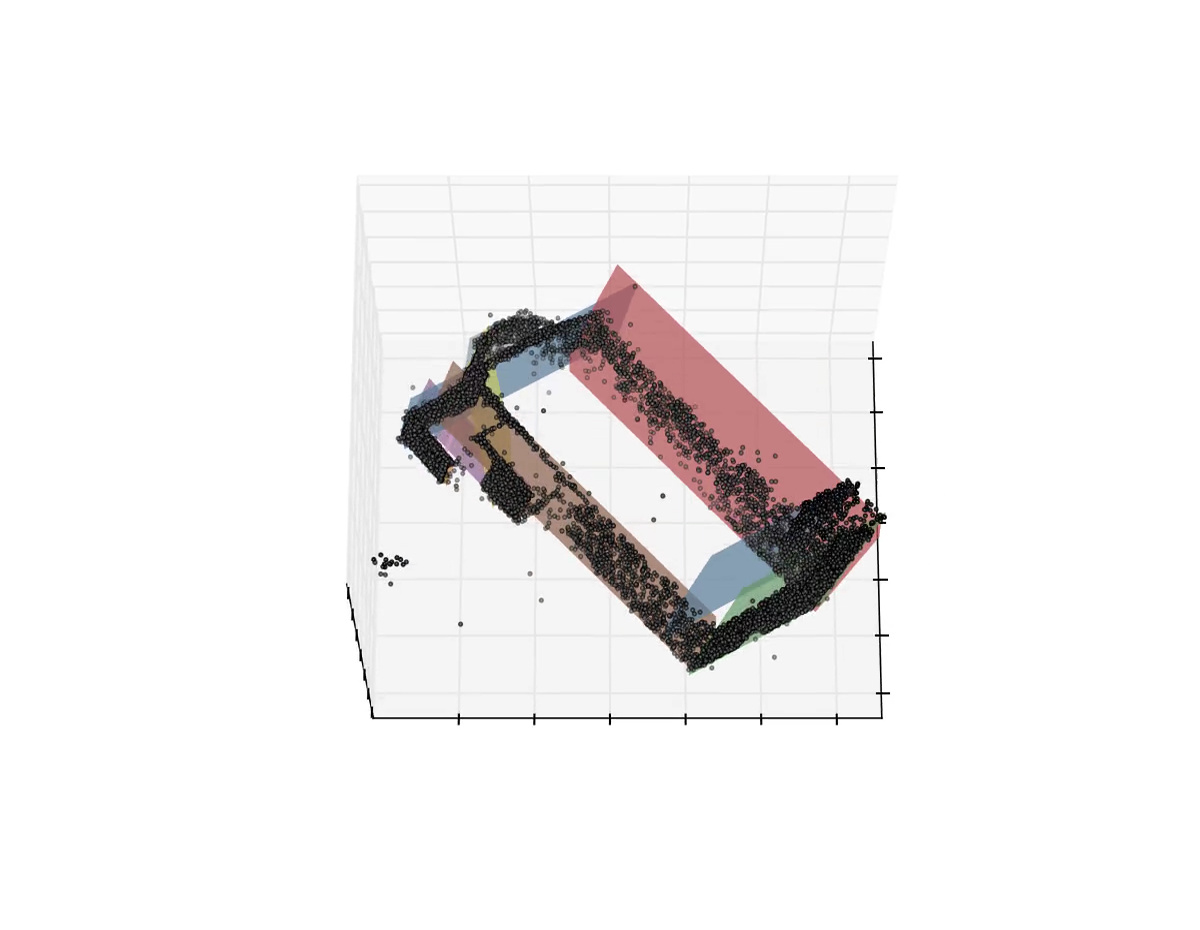} &
			\includegraphics[width=\linewidth,trim={250px 180px 210px 100px},clip]{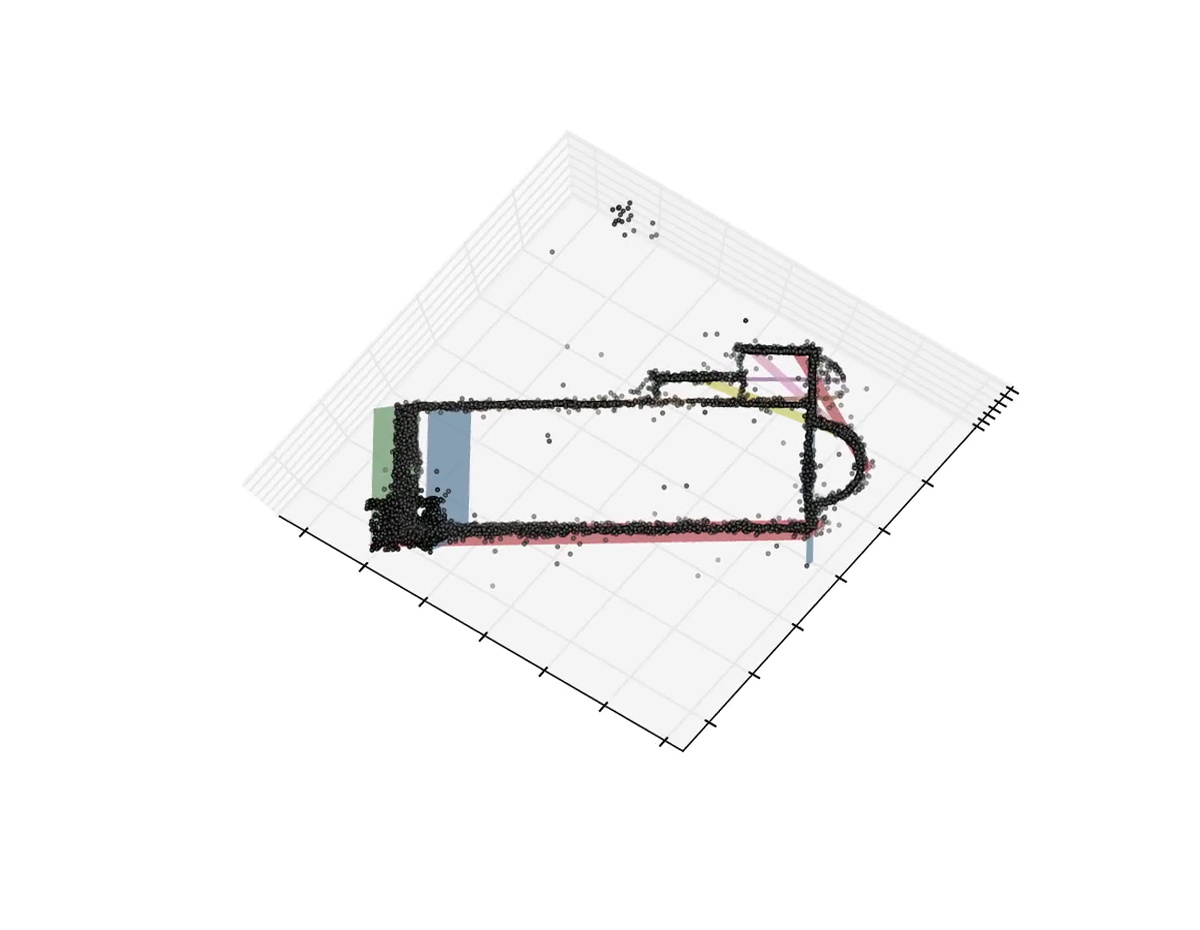} &
			\includegraphics[width=\linewidth,trim={250px 180px 210px 100px},clip]{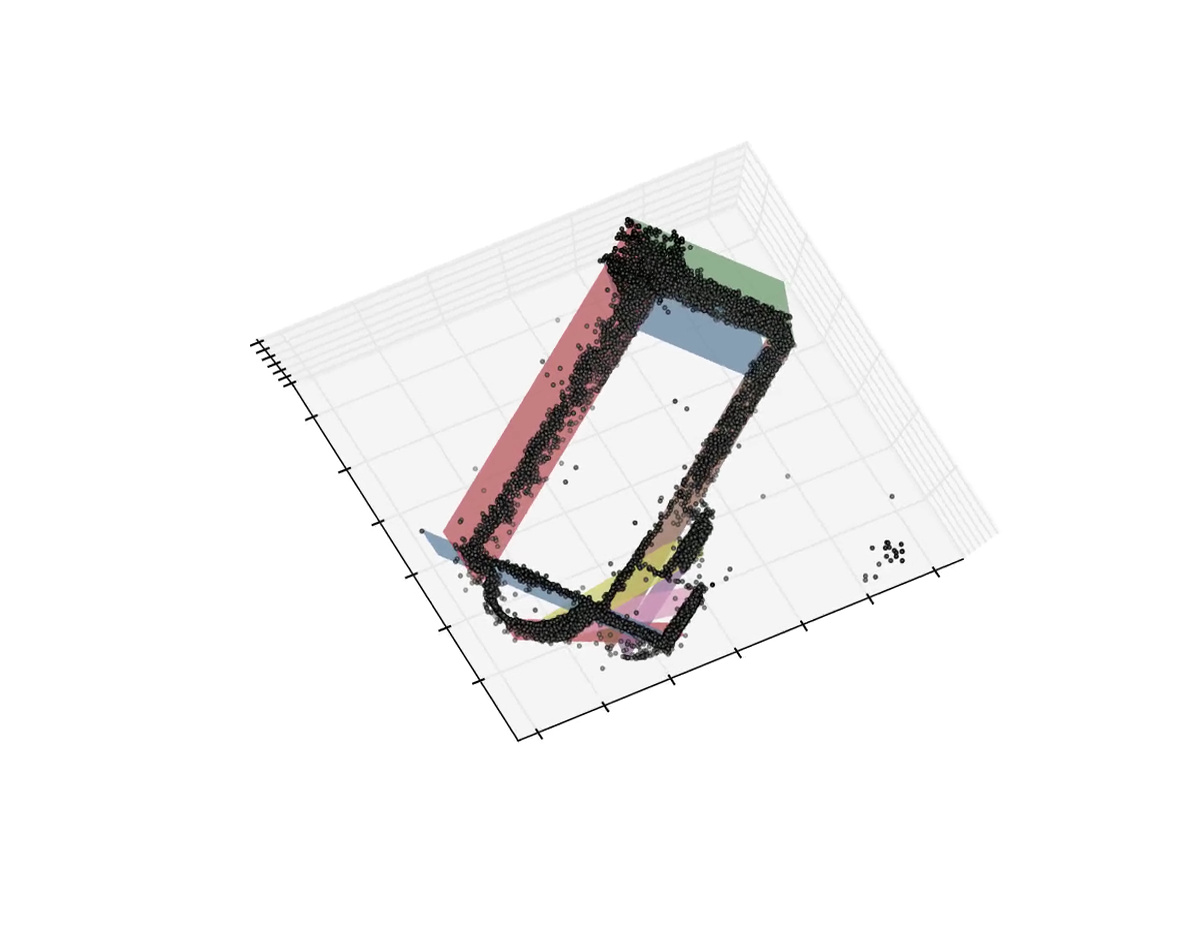} \\
						
			\includegraphics[width=\linewidth,trim={250px 180px 250px 100px},clip]{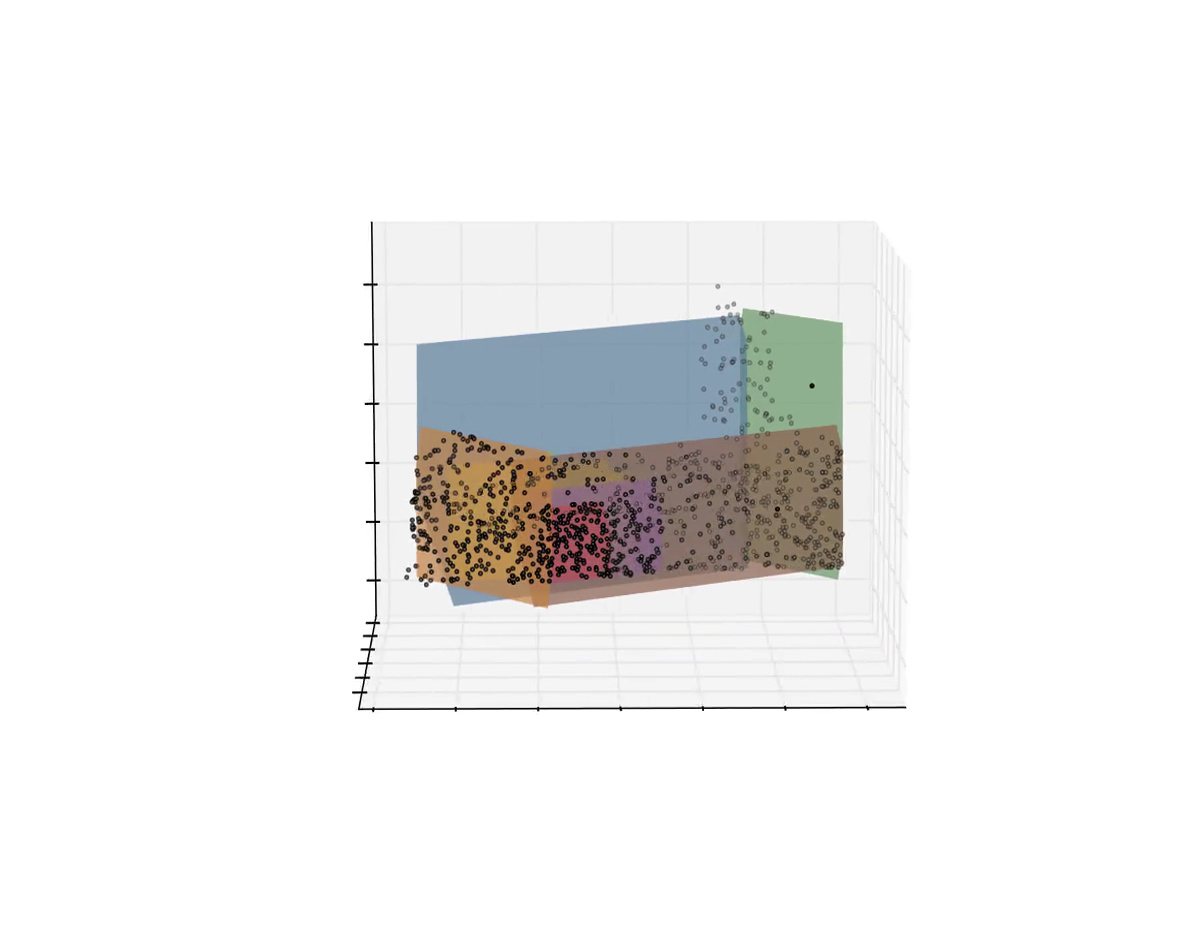} &
			\includegraphics[width=\linewidth,trim={250px 180px 250px 100px},clip]{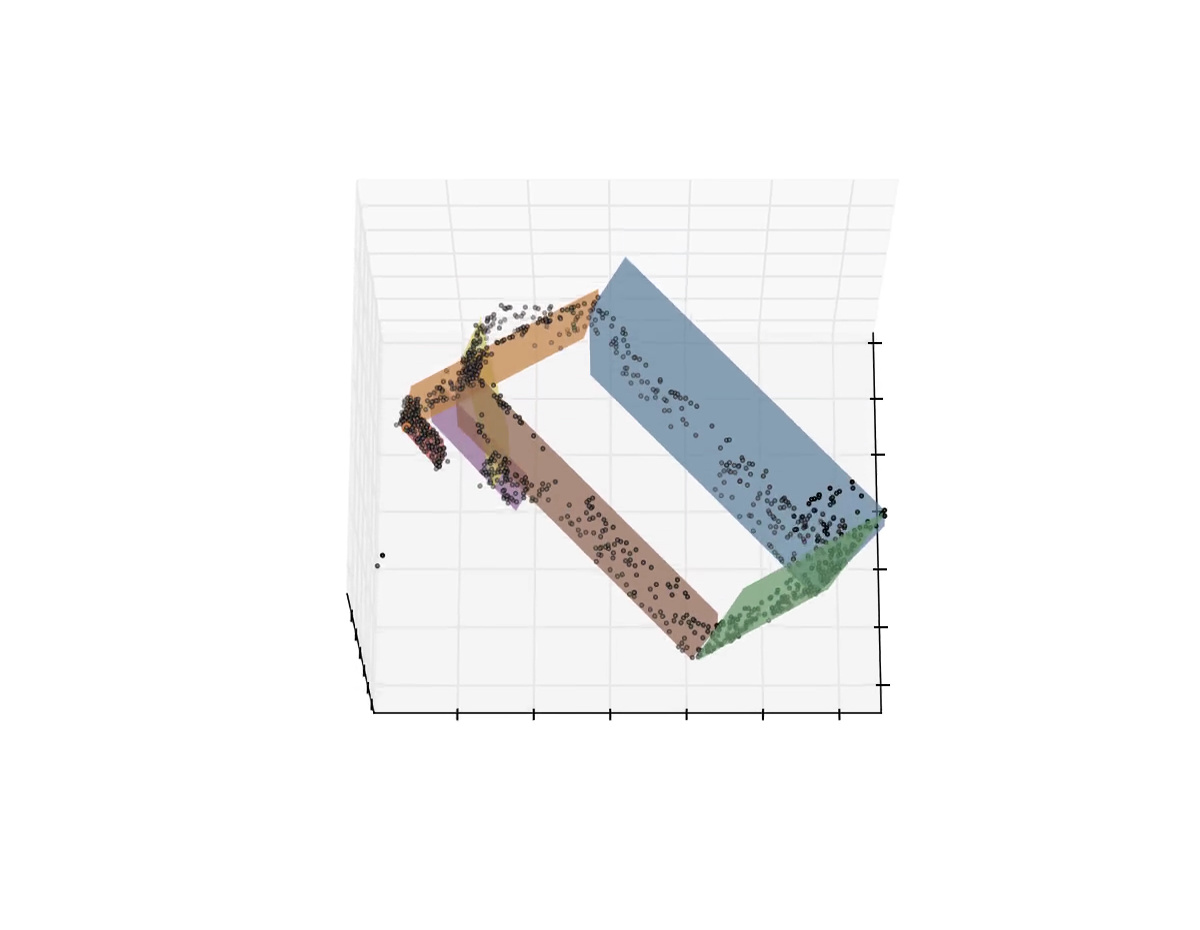} &
			\includegraphics[width=\linewidth,trim={250px 180px 210px 100px},clip]{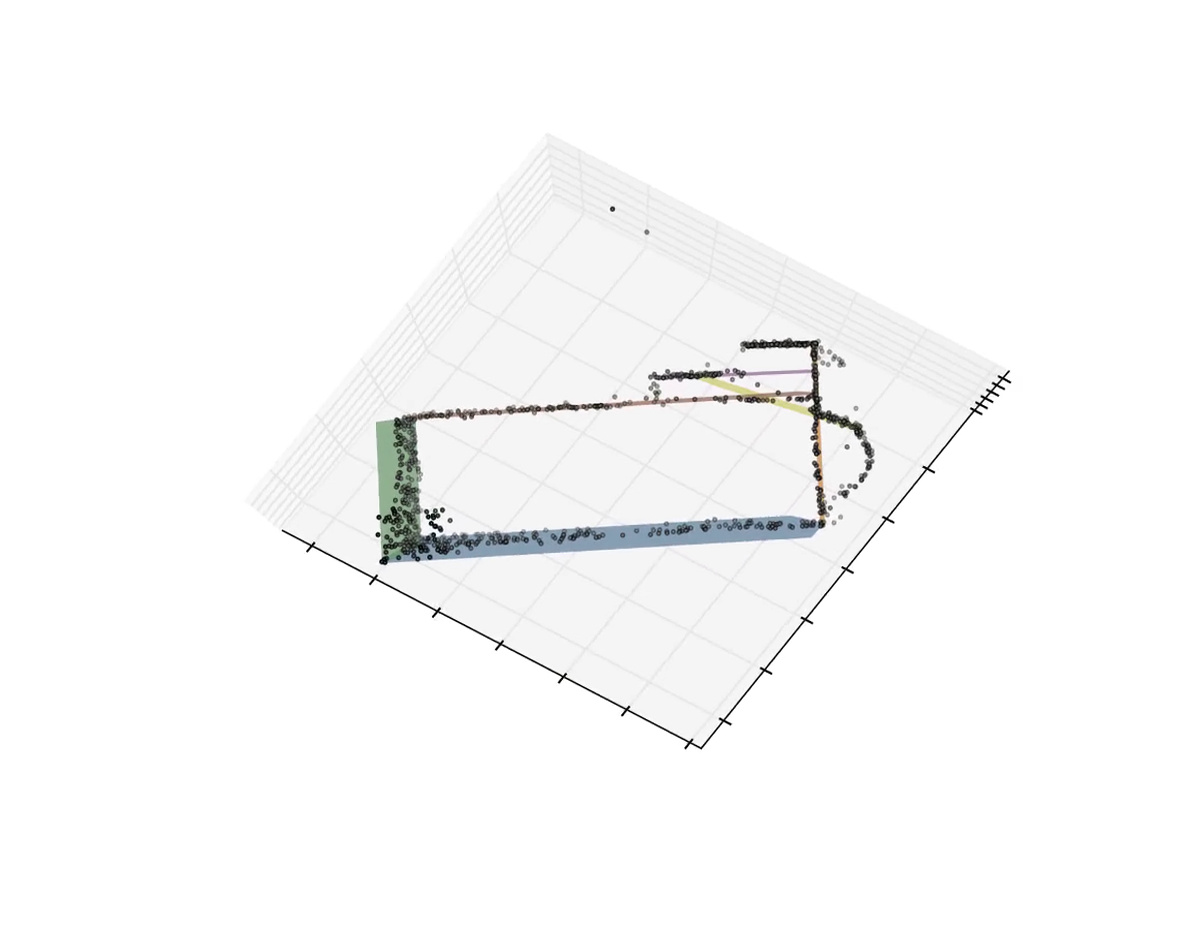} &
			\includegraphics[width=\linewidth,trim={250px 180px 210px 100px},clip]{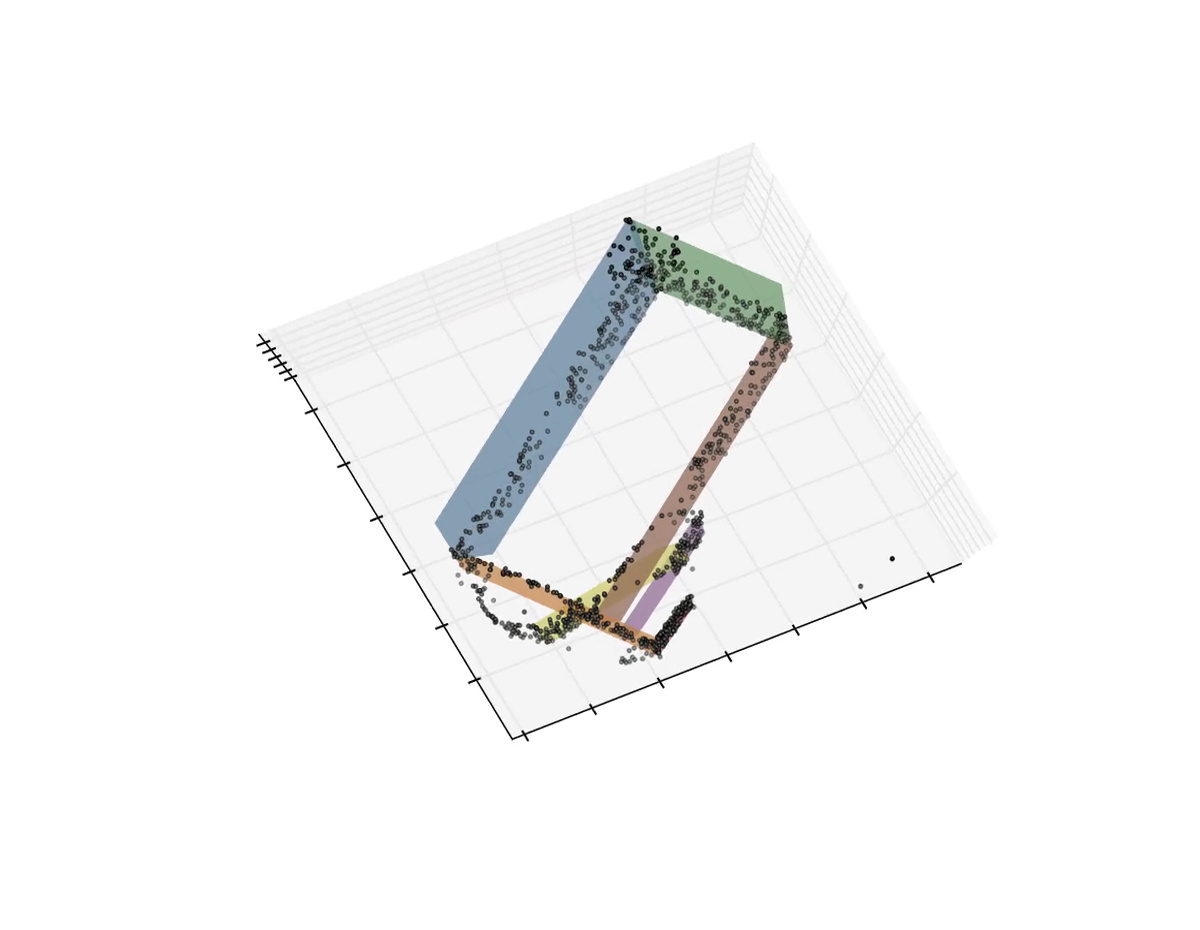} \\
		\end{tabu}

		\caption{The planes estimated by ARSE seen from different angles. We show on the top row the results using the original dataset ($10875$ points) and the dataset with 10x subsampling on the bottom row ($1088$ points).}
		\label{fig:pozzoveggiani_models}		
	\end{subfigure}

	\begin{subfigure}{\textwidth}
		\includegraphics[width=.48\textwidth]{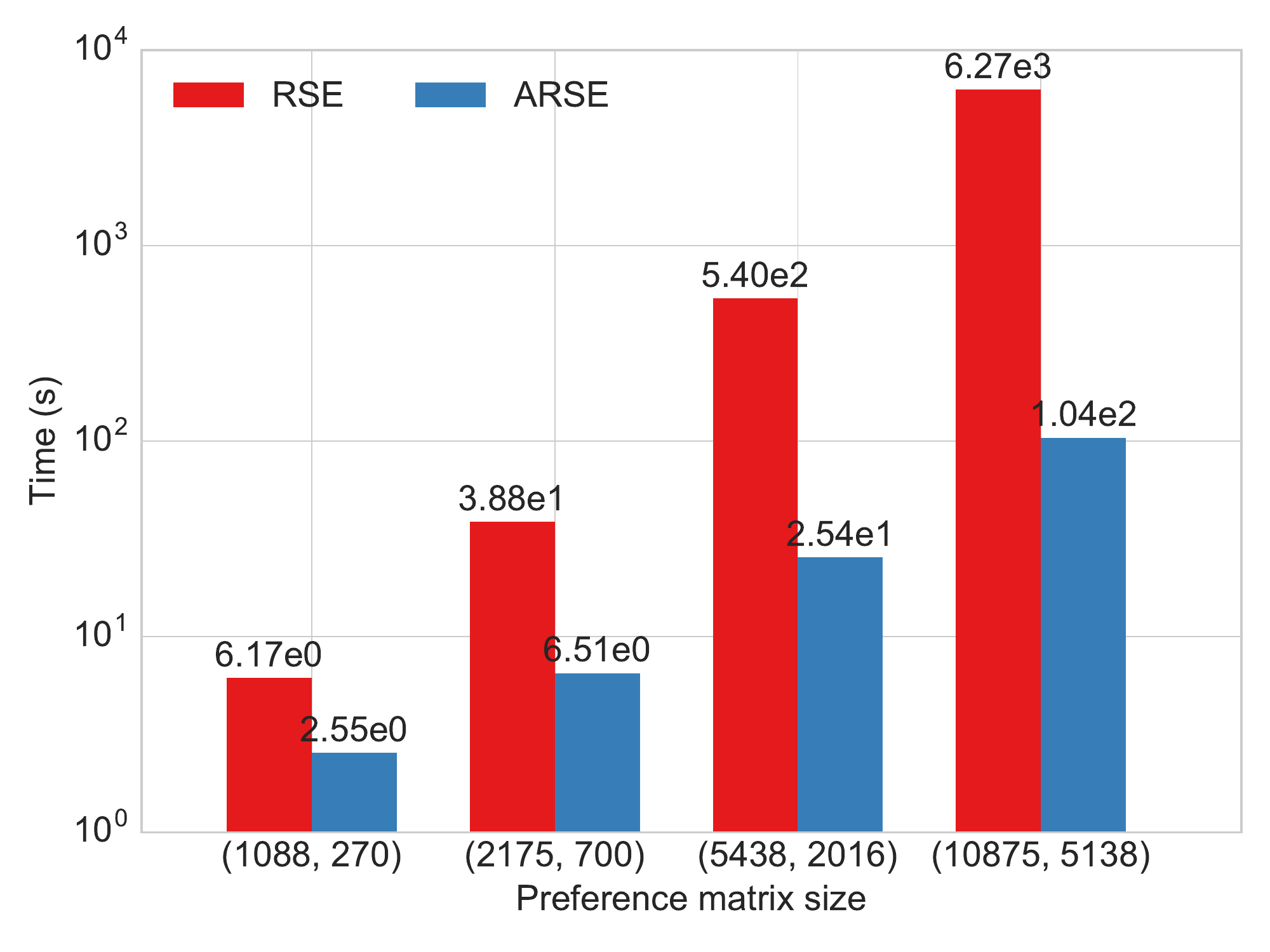}%
		\hfill
		\includegraphics[width=.48\textwidth]{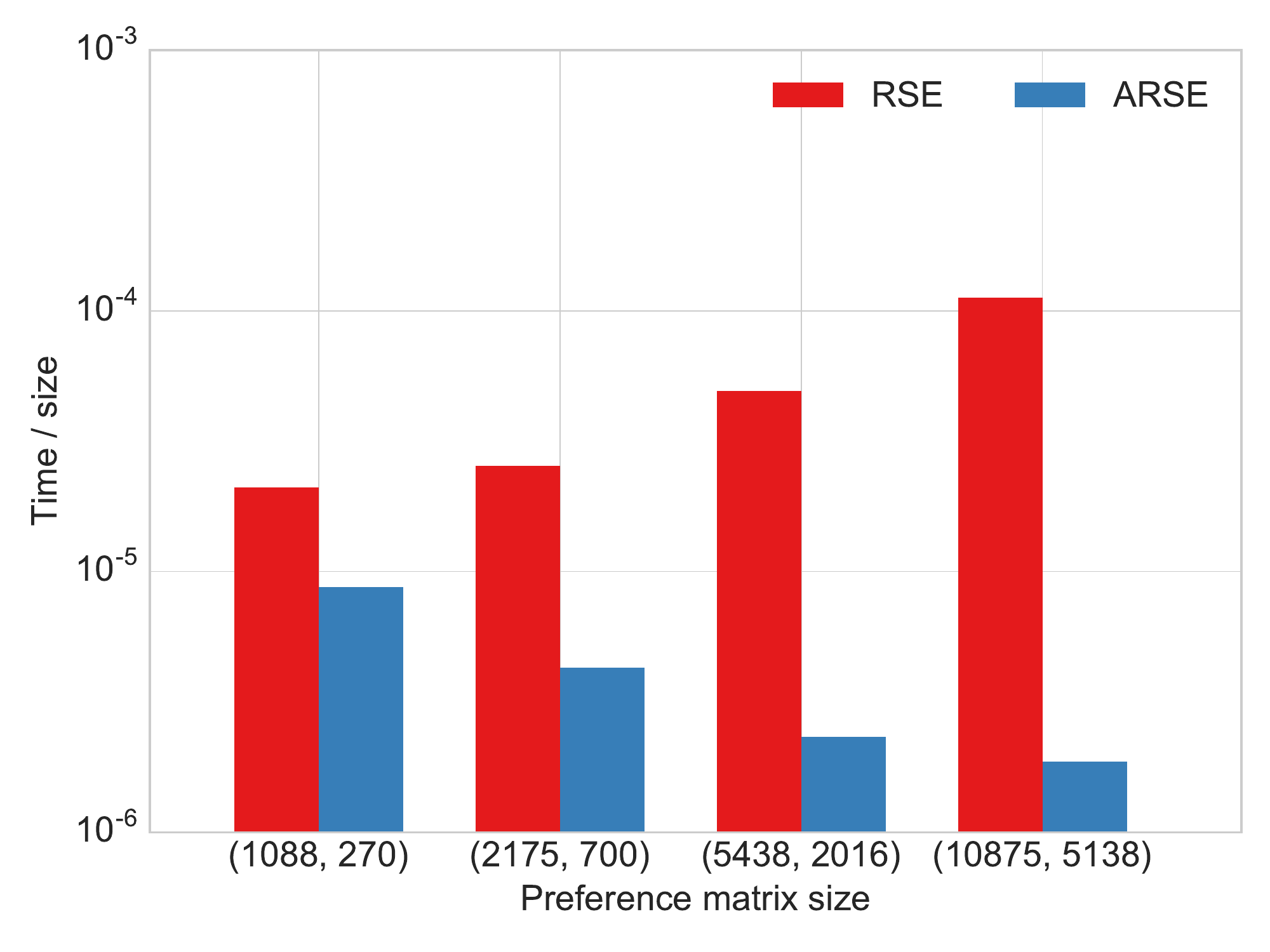}%
		
		\caption{On the left, comparison between the running times of RSE and ARSE as the dataset size (and hence the preference matrix size) increase. As this progression continues, RSE becomes much faster than ARSE (notice the logarithmic scale). On the right, we plot the run-time dividided by the preference matrix size, giving a common ground to compare the performance for different preference matrix sizes. An algorithm linear in the preference matrix size would exhibit a flat behavior in this plot; we can easily observe that while RSE is supralinear, ARSE is clearly sublinear.}
		\label{fig:pozzoveggiani_times}
	\end{subfigure}		
	
	\caption{3D planes estimation. We run the whole algorithmic pipeline on four different versions of the Pozzoveggiani dataset, created by subsampling the original dataset $10, 5, 2, 1$ times, yielding $1088$, $2175$, $5438$, and $10875$ 3D points, respectively. For this experiment, we set $\delta = 10^{-1}$ (\cref{eq:consensusSet}).}
	\label{fig:pozzoveggiani}
\end{figure}

As an example, we show the capabilities of our method in this setting with the detection of 3D planes on the Pozzovegianni dataset.\footnote{\label{note:planes_dataset}\url{http://www.diegm.uniud.it/fusiello/demo/samantha/}}
In this case, the 3D points are obtained from different images of a building (\cref{fig:pozzoveggiani_images}) with a sparse multi-view 3D reconstruction algorithm. Our method recovers the 3D planes in the scene and properly reconstructs the building structure. We evaluate our results on four different versions of the dataset, using $10\%$, $20\%$, $50\%$, and $100\%$ the points. In \cref{fig:pozzoveggiani_models}, we show the estimated planes on the first ($10\%$) and last ($100\%$) versions. Both results are very similar except for the bell tower, which is correctly recovered in the latter case but not in the former (mainly because there are not enough points for a proper estimation).

The experimental setup used in the above paragraph provides a good opportunity to measure the speed difference between RSE and ARSE, see \cref{fig:pozzoveggiani_times}. ARSE is faster than RSE for small datasets, as in all the previous examples; however, the speed difference is not as striking ($2.4$ times faster when using $10\%$ of the points). When working with increasingly bigger datasets, this difference becomes more and more pronounced ($60$ times faster when using $100\%$ of the points), see the left plot in \cref{fig:pozzoveggiani_times}. The other appealing quality of ARSE is that its speed scales very gracefully with the preference matrix size, see the right plot in \cref{fig:pozzoveggiani_times}.

\begin{figure}[t]
	\centering
	
	\includegraphics[width=.48\textwidth]{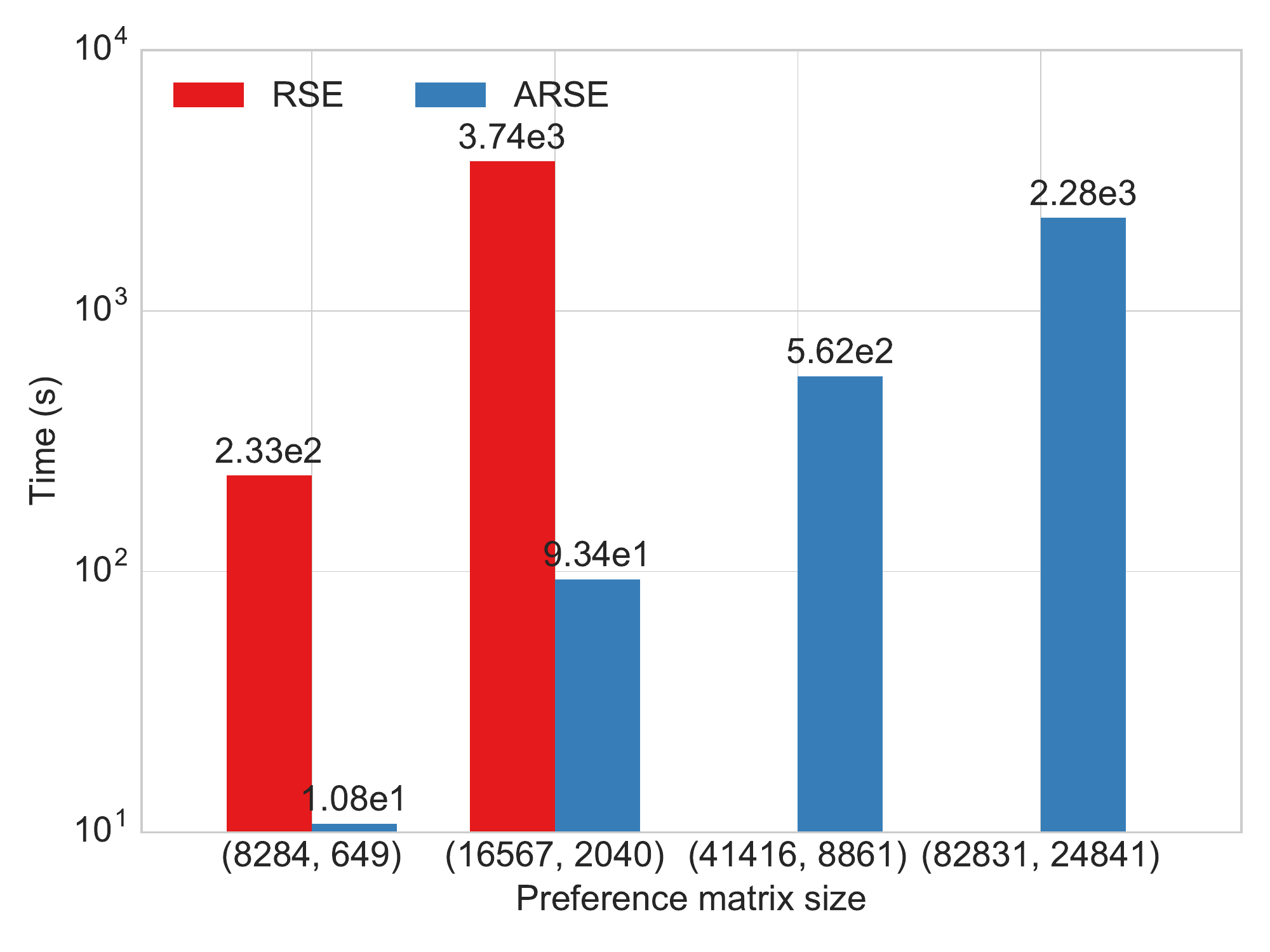}%
	\hfill
	\includegraphics[width=.48\textwidth]{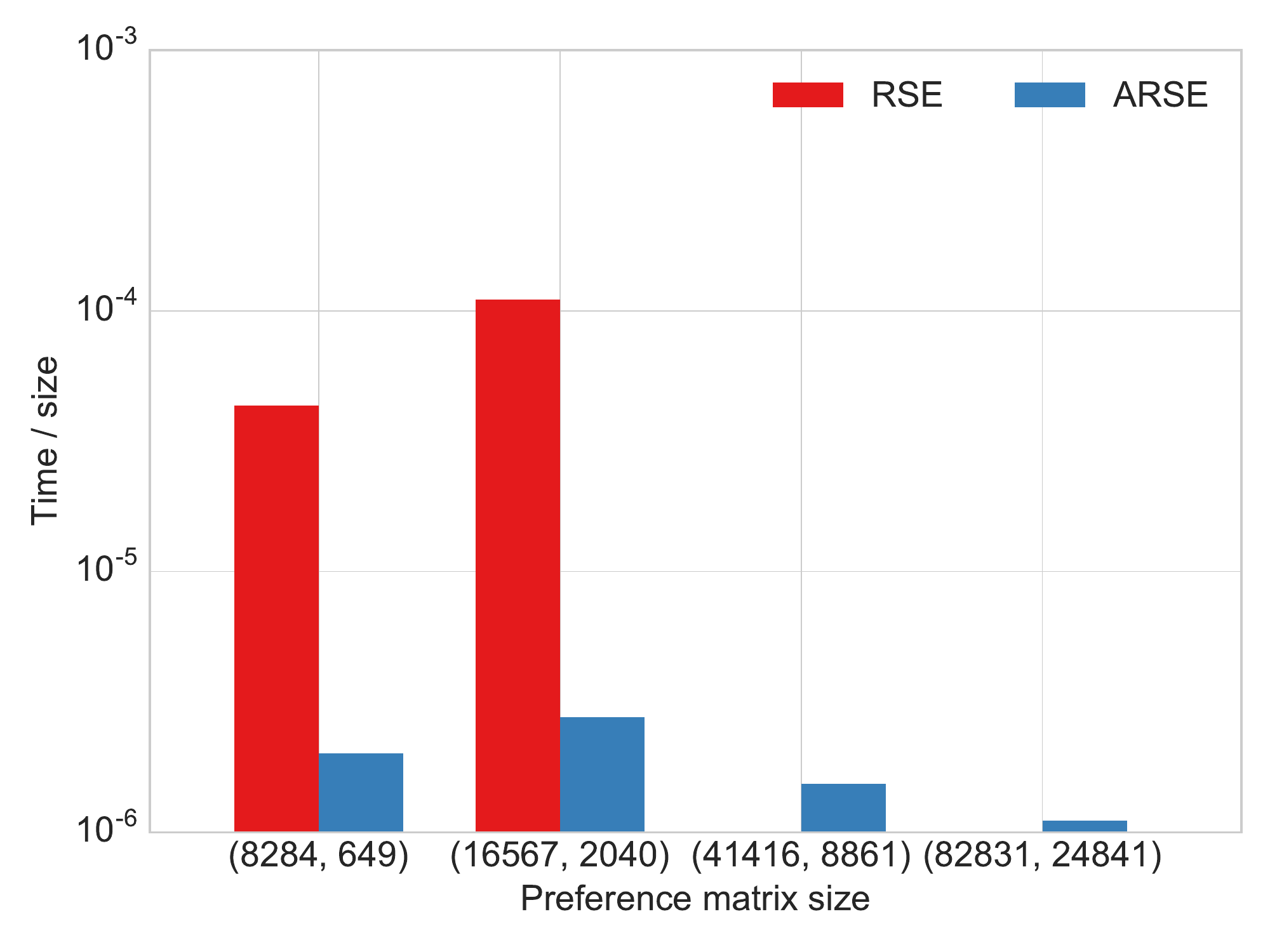}%
	
	\caption{3D planes estimation. Same experiment as in \cref{fig:pozzoveggiani} with the Piazza Bra dataset, created by subsampling the original dataset $10, 5, 2, 1$ times, yielding $8284$, $16567$, $41414$, and $82831$ 3D points, respectively. For this experiment, we set $\delta = 10^{-1}$ (\cref{eq:consensusSet}).
	}
	\label{fig:piazza_bra_times}
\end{figure}

We ran the same experiment on the Piazza Bra dataset,\footnote{See \cref{note:planes_dataset}} see \cref{fig:piazza_bra_times}. We only ran RSE on the dataset versions that include $10\%$ and $20\%$ of the points, because of RAM memory issues. Notice that ARSE with a $82831 \times 24841$ preference matrix is faster than RSE with a $16567 \times 2040$ preference matrix (the former matrix being about 61 times bigger than the latter).

\section{Conclusions}
\label{sec:conclusions}

In this work we introduced a complete and comprehensive algorithmic pipeline for multiple parametric model estimation.
The proposed approach takes the information produced by a random sampling algorithm (e.g., RANSAC) and analyzes it from a machine learning/optimization perspective, using a \textit{parameterless} biclustering algorithm based on L1 nonnegative matrix factorization (L1-NMF).

This new formulation conceptually changes the way that the data produced by the popular RANSAC, or related model-candidate generation techniques, is analyzed.
It exploits consistencies that naturally arise during the RANSAC execution, while explicitly avoiding spurious inconsistencies.
Additionally and contrarily to the main trends in the literature, the proposed modeling does not impose non-intersecting parametric models.

We also presented an accelerated version of the biclustering process, introducing an accelerated algorithm to solve L1-NMF problems. This allows to solve medium-sized problems faster while also extending the usability of the algorithm to much larger datasets. We emphasize that this contribution exceeds the context of this work, as this accelerated algorithm has potential applications in any other context where an L1-NMF is needed.

As a future line of research, we are currently investigating whether using a hard thresholding scheme is actually necessary. Instead of working with binary data, we could work with a real-valued object-model distance matrix, eliminating a critical parameter that has been haunting the RANSAC framework for years.

\bibliography{arse}

\end{document}